\documentclass[journal,twocolumn,transmag, review = true]{IEEEtran}
\usepackage{caption}
\usepackage{times}
\usepackage{epsfig}
\usepackage{graphicx}
\usepackage{amsmath}
\usepackage{amssymb}
\usepackage{multirow}
\usepackage{subfigure}
\usepackage{ulem}
\usepackage{color} 
\usepackage{enumitem}

\usepackage[backref]{hyperref}
\usepackage{makecell}

\usepackage{tikz}
\newcommand*{\circled}[1]{\lower.7ex\hbox{\tikz\draw (0pt, 0pt)%
		circle (.5em) node {\makebox[1em][c]{\small #1}};}}

\ifCLASSINFOpdf
\else
\fi
\begin{document}
%
\title{Multigranular Visual-Semantic Embedding for Cloth-Changing Person Re-identification}


\author{Zan Gao, Member, IEEE, Hongwei Wei, Weili Guan, \\ Weizhi Nie, Member, IEEE, Meng Liu, Member, IEEE, and Meng Wang, IEEE Fellow}

\markboth{Journal of Latex}%
{Shell \MakeLowercase{\textit{et al.}}: Bare Demo of IEEEtran.cls for IEEE Transactions on Magnetics Journals}
%




\maketitle
\begin{abstract}
Abstract - Person reidentification (ReID) is a very hot research topic in machine learning and computer vision, and many person ReID approaches have been proposed; however, most of these methods assume that the same person has the same clothes within a short time interval, and thus their visual appearance must be similar. However, in an actual surveillance environment, a given person has a great probability of changing clothes after a long time span, and they also often take different personal belongings with them. When the existing person ReID methods are applied in this type of case, almost all of them fail. To date, only a few works have focused on the cloth-changing person ReID task, but since it is very difficult to extract generalized and robust features for representing people with different clothes, their performances need to be improved. Moreover, visual-semantic information is often ignored. To solve these issues, in this work, a novel multigranular visual-semantic embedding algorithm (MVSE) is proposed for cloth-changing person ReID, where visual semantic information and human attributes are embedded into the network, and the generalized features of human appearance can be well learned to effectively solve the problem of clothing changes. Specifically, to fully represent a person with clothing changes, a multigranular feature representation scheme (MGR) is employed to focus on the unchanged part of the human, and then a cloth desensitization network (CDN) is designed to improve the feature robustness of the approach for the person with different clothing, where different high-level human attributes are fully utilized. Moreover, to further solve the issue of pose changes and occlusion under different camera perspectives, a partially semantically aligned network (PSA) is proposed to obtain the visual-semantic information that is used to align the human attributes. Most importantly, these three modules are jointly explored in a unified framework. Extensive experimental results on four cloth-changing person ReID datasets demonstrate that the MVSE algorithm can extract highly robust feature representations of cloth-changing persons, and it can outperform state-of-the-art cloth-changing person ReID approaches. Compared with DenseNet121, it can achieve improvements of 45.3\% (21.3\%), 13.0\% (-), 1.5\% (3.4\%), and 3.2\% (1.8\%) on the LTCC, PRCC, Celeb-reID, and NKUP datasets in terms of rank-1 (mAP), respectively\footnote{Manuscript received Feb-22th, 2021; This work was supported in part by the National Natural Science Foundation of China (No.61872270, No.62020106004, No.92048301, No.61572357).  Young creative team in universities of Shandong Province (No.2020KJN012), Jinan 20 projects in universities (2020GXRC040). New Artificial Intelligence project towards the integration of education and industry in Qilu University of Technology (No.2020KJC-JC01). Tianjin New Generation Artificial Intelligence Major Program (No.18ZXZNGX00150, No.19ZXZNGX00110).

Z. Gao and H.W Wei (Corresponding Author) are with the Shandong Artificial Intelligence Institute, Qilu University of Technology (Shandong Academy of Sciences), Jinan, 250014, P.R China. Z. Gao is also with the Key Laboratory of Computer Vision and System, Ministry of Education, Tianjin University of Technology, Tianjin, 300384, China.

Weili Guan is with the Faculty of Information Technology, Monash University Clayton Campus, Australia.  

W.Z Nie is with the School of Electrical and Information Engineering, Tianjin University, Tianjin 300072, China. 

M. Liu is with the School of Computer Science and Technology, Shandong Jianzhu University, Jinan, 250101, China.  

M. Wang is with the school of Computer Science and Information Engineering, Hefei University of Technology, Hefei, 230009, P.R China. 
}.
 
\end{abstract}


\maketitle

\IEEEdisplaynontitleabstractindextext

%
\IEEEpeerreviewmaketitle

\begin{figure}[t]
\begin{center}
\includegraphics[width=3.5in,height = 2in]{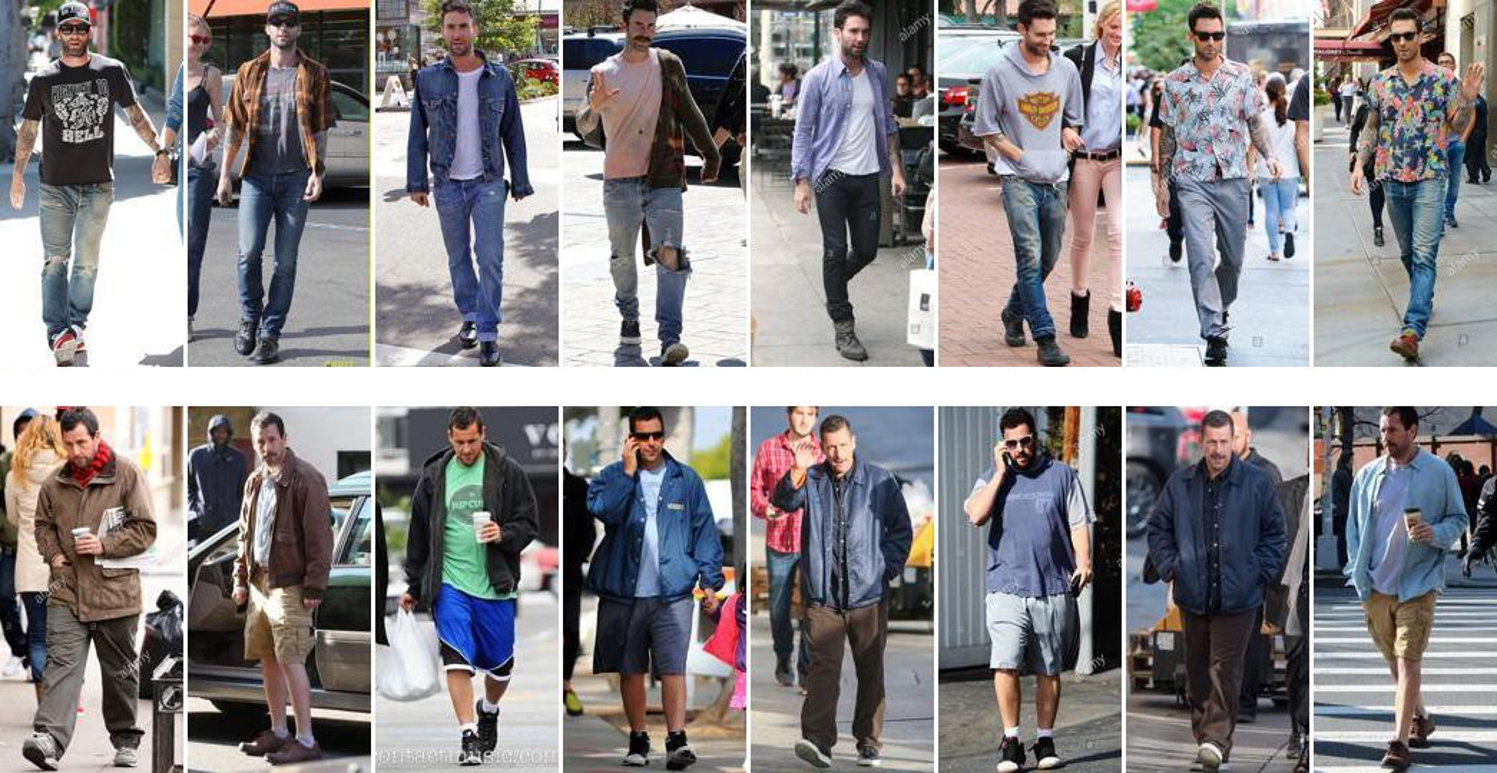}
\caption{Examples of cloth-changing person ReID images. In each row, these images belong to the same person with different clothes.}
\end{center}
\end{figure}

\section{Introduction}
\IEEEPARstart
Person reidentification (ReID) aims to solve the problem of retrieving a target pedestrian across nonoverlapping cameras, where the system uses computer vision technology to determine whether there is a specific person in the images or video sequences taken by different cameras \cite{zheng2015scalable, zheng2017unlabeled}. Person ReID can be combined with person detection and person tracking technology; thus, it plays important roles in urban planning, intelligent monitoring, and security monitoring (e.g., looking for rioting demonstrators at the U.S. Capitol). Since lighting variations and changes in poses or observation perspectives often occur, and the visual appearances of pedestrians change drastically, thus, this is a very challenging and difficult task. To date, many person ReID approaches have been proposed \cite{Wang2016Zero, zheng2019pedestrian, Gao2020TSA, Lin2019Improving, Zhang2021Person, Luo2020STN, Yang2020Spatial, Wei2019GLAD, sun2018beyond, Sun2021Learning, Zhou2018Large, wang2018learning, Gao2020DCR}, and some person ReID datasets, including Market1501 \cite{zheng2015scalable}, DukeMTMC-reID \cite{zheng2017unlabeled}, and CUHK03 \cite{li2014deepreid}, have been released, but they often assume that the same person viewed by nonoverlapping cameras has the same clothes and that their visual appearance is similar. To some extent, this assumption is true; in a short time frame, a specific person is likely to appear in multiple nonoverlapping camera perspectives without changing clothes; thus, the appearance of clothes can be used as an identifying feature. Moreover, this assumption also simplified the task objectives in early studies and promoted their development. However, in an actual surveillance environment, a person has a great probability of changing clothes after a long time span; furthermore, they also often take different personal belongings with them, or a criminal suspect deliberately changes his/her dress to avoid detection; thus, the visual appearance feature changes dramatically. Fig. 1 shows some examples of cloth-changing person ReID images. From them, we can observe that the differences in the visual appearances of the same person with different clothes are very large, and it is also very difficult for humans to identify them. If the existing person ReID approaches are directly applied in this case, their performances deteriorate dramatically, and they often fail. Thus, to solve this issue, one must urgently solve the cloth-changing person ReID task, which is a more challenging and practical task.

To date, few works have focused on the cloth-changing person ReID task. Previously, researchers \cite{sun2018beyond, wang2018learning} paid more attention to local features or partial features. For example, Sun et al. \cite{sun2018beyond} proposed a PCB module to extract fine-grained information based on partial human images; Wang et al. \cite{wang2018learning} proposed a multibranch network architecture to jointly learn global feature representations and local feature representations. However, these methods are not specially designed for the cloth-changing person ReID task, and their performances are unsatisfactory on this task. Thus, few researchers \cite{huang2020beyond, yang2020clothing, qian2020long, zheng2019joint, yu2020cocas} have tried to specially design different approaches for the cloth-changing person ReID task; for example, Huang et al. \cite{huang2020beyond} proposed a vector neuron concept where the direction of the vector was used to represent a change in clothing information, and the length of the vector was utilized to identify the person. In \cite{yang2020clothing}, human contour sketching information was used as a substitute for human color information; in this way, visual appearance changes caused by changes in clothing can be reduced. Qian et al. \cite{qian2020long} proposed a shape embedding module and a cloth-eliminating shape-distillation module. The former is used to encode shape information from human body keypoints, and the latter is utilized to adaptively distill the identity-relevant shape feature and explicitly disentangle the identity-irrelevant clothing information. Zheng et al. \cite{zheng2019joint} used a generative model to automatically generate person images with different clothing appearances. Yu et al. \cite{yu2020cocas} proposed a new solution for changing clothes where rich clothing templates were supplied, and then the target person wearing these different clothing templates was also used as the query. These approaches are very good for trying to solve the cloth-changing person ReID problem, but their extracted features are not robust and still related to the clothes; thus, their performances are unsatisfactory. Moreover, visual semantic information and human attributes are often ignored, or their roles in cloth-changing person ReID are not fully explored. However, this information, such as human attributes, is often invariant for clothing changes; thus, it can supply important clues.

To solve these issues, in this work, we propose a novel end-to-end MVSE algorithm for cloth-changing person ReID, where the key idea is to shield clues related to the appearance of clothes and only focus on high-level human attribute information that is not sensitive to view/posture changes. Specifically, a CDN module is designed, where multiple parallel convolutional layers are used to obtain different clothing appearance features for the same identity, and then vectors are utilized to represent different high-level human attributes. Then, a PSA module is proposed to obtain the human semantic information, and it can be used to align the human attributes. Moreover, an MGR scheme is employed to focus on the unchanged portions of humans. Extensive experimental results on four cloth-changing person ReID datasets demonstrate that MVSE can outperform state-of-the-art person ReID approaches.

The main contributions of this paper are summarized as follows.
\begin{itemize}[leftmargin=*]
\item We develop a novel MVSE algorithm for cloth-changing person ReID, where visual semantic information and human attributes are embedded into a network, and they are jointly optimized in a unified framework. MVSE shields clues related to the appearance of clothes and only focuses on high-level human semantic information that is not sensitive to view/posture changes.

\item We design a CDN module to improve the feature robustness of the algorithm for a person with different clothes, where different high-level human attributes are fully utilized. Then, we develop a PSA module to obtain visual-semantic information and align the human attributes. Moreover, the MGR scheme is employed to focus on the unchanged parts of the human. In this way, a generalized and robust appearance feature can be extracted, and it can effectively solve the problem of clothing changes and the issues of pose changes and occlusion under different camera perspectives.

\item We first systematically and comprehensively evaluate the MVSE algorithm on four public cloth-changing person ReID datasets, and the experimental results demonstrate that the MVSE approach can outperform state-of-the-art cloth-changing person ReID methods in terms of the mAP and rank-1.

\end{itemize}

The remainder of the paper is organized as follows. Section II introduces the related work, and Section III describes the proposed MVSE method. Section IV describes the experimental settings and the analysis of the results. Section V presents the details of the ablation study, and concluding remarks are presented in Section VI.

\section{Related Work}

Person ReID is a hot research topic in machine learning and computer vision, and it is also widely used in video surveillance; thus, many person ReID approaches have been proposed. According to the person's visual appearance, these methods can be roughly divided into clothing-consistent person ReID and cloth-changing person ReID. In the following, we will separately introduce them.

\subsection{Clothing-Consistent Person ReID}
As mentioned earlier, almost all existing ReID datasets \cite{zheng2015scalable, zheng2017unlabeled, li2014deepreid} are captured over a short time period. As a result, for the same person, the visual appearance of the clothes is consistent. By learning discriminative features \cite{yao2019deep, zheng2019joint} or robust distance metrics \cite{paisitkriangkrai2015learning, shen2018end}, people have made efforts to develop related methods for clothing-consistent person ReID. For example, Liu et al. \cite{liu2017hydraplus} proposed an HP-Net module where multilayer attention mechanism graphs were mapped to different feature layers in multiple directions. The HP-Net module can capture attention from the shallow layer to the semantic layer, mine multi-scale optional attention features, and enrich the final pedestrian feature representation. In \cite{song2018mask}, a body mask was used as an additional input and accompanied by an RGB image to enhance the ReID feature learning process; this can help reduce background clutter, while also including identity-related functions, such as body shape information. Miao et al. \cite{miao2019pose} proposed a pose estimator to detect the key points of the human body from a pedestrian image, and then a threshold was set to screen whether there was occlusion in each area. Sun et al. \cite{sun2018beyond} proposed a part-based convolutional baseline (PCB) module and then used the local feature representation method of horizontal segmentation to merge the local features of horizontal tiles to obtain a powerful ReID feature; this approach became an important benchmark for person reidentification problems. Sun et al. \cite{sun2019perceive} first pre-trained a model to recognize human body regions on global person images, and then each region was located, and the regional-level features were extracted. Finally, when calculating the distance between two images, the regional distance and the global distance were jointly calculated. Gao et al. \cite{Gao2020DCR} proposed a DCR algorithm, where a deep spatial pyramid-based 
collaborative feature reconstruction model was built to solve occlusion problems, pose
changes and observation perspective changes. Li et al. \cite{li2017learning} designed an MSCAN module to learn powerful features, where the local contextual knowledge was well captured by stacking multiscale convolutions on each layer. In this way, the learned body parts can solve the issues of pose changes and background clutter in the given feature representation based on parts.
Gao et al. \cite{Gao2020TSA} proposed a novel texture semantic alignment (TSA)
approach with visibility awareness for the partial person ReID task, where the occlusion issue and changes in poses were simultaneously explored in an end-to-end unified framework. Although these models are robust to changes caused by poses, lighting, and viewing angles, they are vulnerable to clothing changes, as the models are heavily reliant on clothing appearance consistency.

\subsection{Cloth-changing Person ReID}
As the cloth-changing person ReID problem is a new research topic and the visual appearance of a person varies greatly over a long time span, it is very difficult to extract generalized and robust appearance features. To date, few works on this topic have been proposed \cite{huang2020beyond, yang2020clothing, zheng2019joint, qian2020long, yu2020cocas}. For example, Huang et al. \cite{huang2020beyond} used vector neurons instead of scalar neurons to design their network. A vector neuron uses the direction to represent changes in clothing information and the length to represent an identity; thus, it can perceive the clothing changes of a specific person. In addition, auxiliary modules can be used to enhance the robustness of the module. Yang et al. \cite{yang2020clothing} proposed a PRCC module, where the auxiliary information of a human body contour sketch was used instead of color information to extract the most adaptable clothing replacement feature. Moreover, a spatial polar transformation (SPT) layer was designed to transform the contour sketch image, and then a multi-stream network was used to aggregate multiple granular features to better discriminate people by changing the sampling range of the SPT layer. Qian et al. \cite{qian2020long} proposed an LTCC module, where the main idea was to completely delete information related to the appearance of clothes and only focus on body shape information that is not sensitive to changes in perspective and posture. In the LTCC, a shape embedding (SE) module encodes the shape information from human body keypoints, and the ‘Cloth-Elimination Shape-Distillation’ (CESD) module employs keypoint embedding to adaptively distill identity-relevant shape features and explicitly disentangle the identity-irrelevant clothing information. Zheng et al. \cite{zheng2019joint} proposed a DG-Net module, where a generative model was utilized to automatically generate person images with different appearances regarding clothing. Yu et al. \cite{yu2020cocas} proposed a new solution for changing clothes called COCAS, where rich clothing templates were supplied; thus, in the query, the clothing template image and an image of the target person wearing other clothes were used as dual inputs to search for the target image. In addition, some cloth-changing datasets, such as the Celeb-reID dataset \cite{huang2020beyond}, PRCC dataset \cite{yang2020clothing}, NKUP dataset \cite{wang2020benchmark}, LTCC \cite{qian2020long} dataset, and COCAS \cite{yu2020cocas} dataset, have been released, and there are differences in viewpoint, light, resolution, body posture, season, background, and clothing with respect to the collected images, and the data are diversified for person ReID. However, since it is very difficult to extract generalized and robust features to represent people with different clothes, their performances need to be improved. Moreover, visual-semantic information and human attributes are often ignored. Thus, in this work, we will fully explore the available visual semantic information and human attributes and then extract a generalized and robust feature to represent a person with different clothes.

\section{The Proposed Approach}
Although many person ReID algorithms have been proposed, few works have focused on the cloth-changing person ReID task; moreover, the inherent visual semantic information is often ignored in these existing approaches. To solve these issues, in this work, a novel MVSE algorithm is developed for the cloth-changing person ReID task, where the visual semantic information and high-level human attributes are embedded into the network. It shields clues related to the appearance of clothes as soon as possible and only focuses on body shape information that is not sensitive to view/posture changes. The MVSE network mainly consists of an MGR module, a CDN module, and PSA modules. Specifically, an image is first fed into the backbone network (in our experiments, DenseNet121 is used as the backbone); thus, feature mappings can be obtained. Then, the MGR module is utilized for these feature mappings; moreover, the outputs of the MGR module are fed into the CDN module and the PSA module. Separately, the RGB image is also fed into HRNet \cite{sun2019deep} to obtain human semantic information, which is used to guide the optimization of the PSA module. Moreover, the outputs of the PSA module are fed into the CDN module to align the human semantics. Most importantly, these three modules are jointly explored in a unified framework. In the following, we will introduce them individually.

\begin{figure*}[t]
\begin{center}
\includegraphics[width=7in,height = 3.0in]{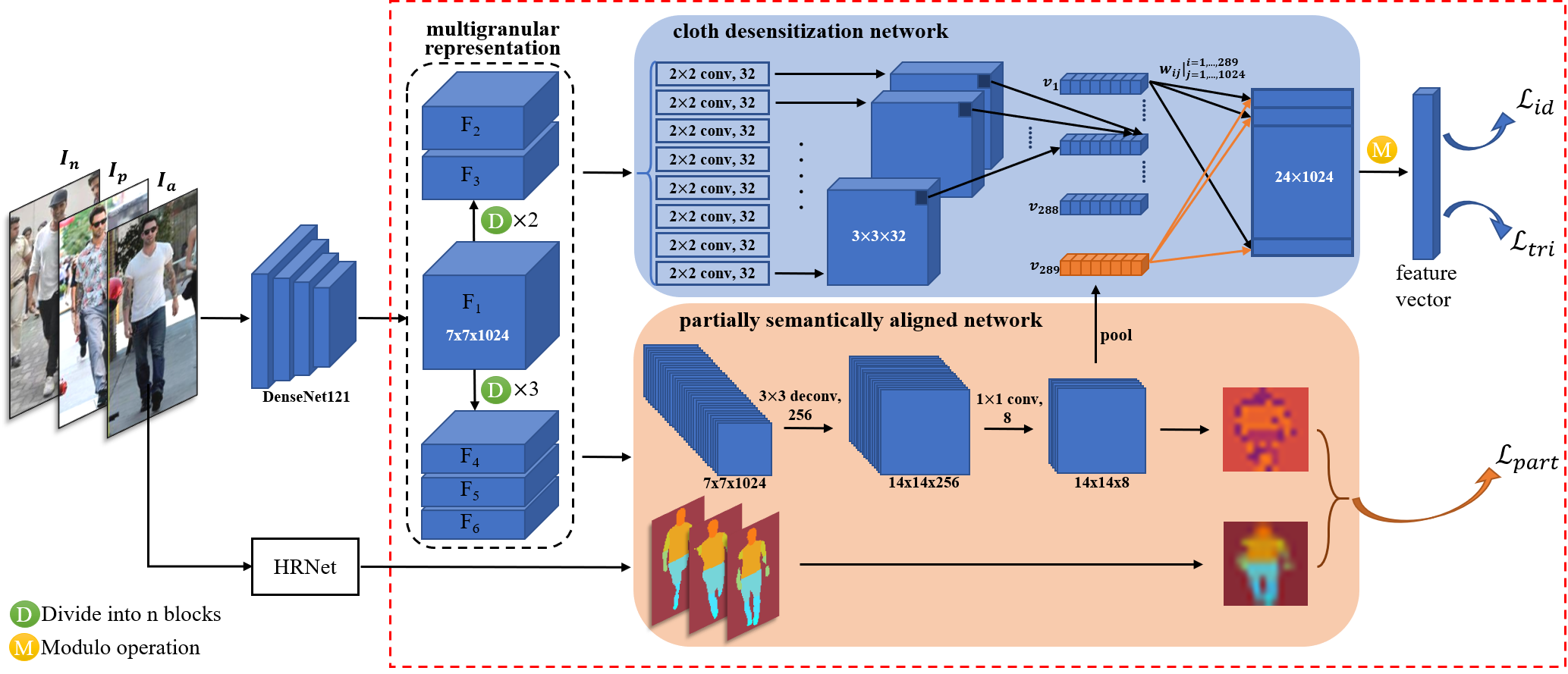}
\caption{Pipeline of the MVSE approach, which mainly consists of the MGR, CDN, and PSA modules. In the MGR module, multigranular features can be learned to focus on the unchanged parts of the human being studied. In the CDN and PSA modules, the inherent high-level human attributes and the visual semantic information are fully explored to improve the generalization and robustness of the proposed method. $D$ indicates the division operation. The parts highlighted by the red dashed lines are our key contributions (best viewed in color).}
\end{center}
\vspace{-1.5em}
\end{figure*}

\subsection{Multigranular Representation (MGR)}

With the development of deep learning techniques \cite{Liu2019Towards, Gao2021Pairwise}, researchers have designed different convolutional neural networks (CNNs) for use in the person ReID task, but since visual appearances are very different in the cloth-changing person ReID task, it is very difficult for an individual feature to extract a generalized and robust feature to represent a person with different clothes. Inspired by the MGN \cite{wang2018learning}, which consists of one branch for the global feature representation and two branches for local feature representations, we propose multigranular representation mechanisms to adaptively extract multilevel and multigranular feature information. The CNN can not only capture the global feature representation of the subject from the entire image but also capture the fine-grained local features of the subject from multiple partial regions. Thus, in the MGR module, different horizontal block operations at the multigranular feature representation layer are performed on $ F \in \mathbb{R}^{7\times7\times1024}$ (the outputs of the backbone) to obtain one complete feature mapping $F_1$, two bisecting feature mappings $ F_2 $ and $ F_3 $, and three trisecting feature mappings $ F_4 $, $ F_5 $, and $ F_6 $. In total, six feature representations of different levels ranging from coarse-grained to fine-grained can be obtained; thus, these multigranular feature representations can compensate for the information loss in the global feature and allow the model to pay increased attention to the unchanged parts of the human. Moreover, these representations are further fed into the CDN and PSA modules. In addition, to improve the feature discrimination ability of the network, an anchor image $I_a$, a positive image $I_p$ from the same person as that in the anchor image, and negative images $I_n$ from different persons are used as anchor images and are simultaneously fed into the backbone (DenseNet121) pretrained by the ImageNet dataset. Thus, the extracted multigranular feature representations can have certain robustness.

\subsection{Clothing Desensitization Network (CDN)}
Although multigranular feature representations are extracted in the MGR module and have certain robustness, since the difference between the visual appearances of a person with different clothes is very large, the generalizability and robustness of the features need to be further improved. Moreover, it is also very difficult for these low-level features to represent people with different clothes. Thus, in this work, we design a CDN module to improve the feature robustness and generalizability for people with different clothes. In the CDN module, different high-level human attributes that are insensitive to cloth-changing issues are fully utilized to describe a person with different clothing.

Specifically, based on the obtained multigranular feature representations, each feature is separately fed into the CDN module, and then a convolutional layer containing eight parallel convolution operations ($2\times 2$ kernel size, stride of 2, and 32 channels) with different weights for the low-level visual feature mappings is utilized to reduce the spatial dimensionality and channel dimensionality of the given feature. Thus, we can obtain eight higher-dimensional feature mappings with different representations; at this time, the dimensionality of each feature map is $ 3 \times 3 \times 32 $. The reason why eight parallel convolution operations are used is that visual appearances are very complex in the cloth-changing person ReID task, and only one convolution network is insufficient for fully representing them; thus, multiple convolution networks can describe the studied person from different perspectives. To obtain human attributes, a reshaping operation is employed to connect the corresponding channels of each feature mapping; thus, $288$ ($3 \times 3 \times 32$) 8-dimensional vectors can be obtained to represent a single attribute of a person. In addition, to align the human attributes, the outputs of the PSA module (we will introduce this module in the next subsection) are also used; thus, in total, $299$ 8-dimensional vectors $v \in R^8$ are used for a single feature representation. Finally, a nonlinear extrusion function is used to standardize the length of each vector, and it can be defined as follows:

\begin{small}
\begin{equation}
\begin{aligned}
& v_k = \frac{||v_k||^2}{1 + ||v_k||^2} \cdot \frac{v_k}{||v_k||} ,
\end{aligned}
\end{equation}
\end{small}

where $v_k$ indicates the $k^{th}$ feature vector of $v$ and $ k \in [1,289] $. $||*||^2$  indicates  the square of the vector norm. To obtain high-level human attributes and make the learned feature more generalized and robust, the weight matrix $w_{ij}$ learned by the network is used for these feature vectors $v$, and the features with high-level human attributes $\bar v$ are related to all $v_k$. It can be computed by

\begin{small}
\begin{equation}
\begin{aligned}
& \bar v = w_{ij} \cdot v ,
\end{aligned}
\end{equation}
\end{small}

where $\bar v \in R^{24 \times 1,024}$ denotes the features with high-level human attributes, $i$ is the index of $v$ ranging from 1 to 289 and $j$ is the index of $\bar v $ ranging from 1 to 1,024.  To further improve the generalizability and robustness of $\bar v$ for a person with different clothes, a coupling calculation needs to be further performed to make it possible to represent each identified person as a corresponding vector; thus, the modulo operation is used as follows:

\begin{small}
\begin{equation}
\begin{aligned}
& C = ||\sum_{k=1}^K u_k \cdot \bar v_k|| ,
\end{aligned}
\end{equation}
\end{small}
where $C$ indicates the outputs of the modulo operation and $u_k$ indicates the coupling coefficients which are learned by the network, and few parameters with large values are kept to activate the high-level human attributes. Note that all of the 8-dimensional vectors are standardized, and then the modulo operation is performed to remove the attribute representation of different clothes; the modulo operation is performed for each dimension of the feature to only express the existence of the clothes attributes, without providing a specific type description. Thus, a 1,024-dimensional feature vector that is generalized and robust for a person with different clothes is used to describe the original input RGB image.

\subsection{Partially Semantically Aligned Network (PSA)}
Since changes in backgrounds, viewpoints and poses or observation perspectives often occur, person images often have the problem of misalignment, and only using high-level human attributes is insufficient for fully describing human appearance. In most of the existing methods \cite{sun2018beyond, sun2019perceive}, the horizontal dicing operation is often used to align and extract fine-grained features, but its disadvantage is that the background and half-body alignment of the pedestrian inevitably lead to negative optimization. In the MGR module, the feature mappings are divided into many regions, and then the combined features from different regions are learned, but we hope these features are different from each other and that there is almost no redundancy. Therefore, a PSA module is designed to obtain the visual-semantic segmentation information and align the human attributes. In the PSA module, we impose some semantic segmentation constraints on the feature mapping, which forces the model to predict the component label from the feature mapping. If the PSA module can predict a part label based on the corresponding feature mapping, its positioning capability will be effectively maintained, thereby reducing redundancy and achieving the purpose of feature alignment. Specifically, in the PSA module, the feature mapping is first passed through a $ 2 \times 2 $ deconvolution layer with a step size of 2; thus, we can obtain a $ 3 \times 3 \times 256 $ feature mapping. Then, the normalization and ReLU activation functions are used. Moreover, a $ 2 \times 2 $ convolutional layer is employed to predict the part labels. Note that the deconvolution layer is used for upsampling, and the $ 2 \times 2 $ convolution layer is used for pixel classification. Thus, we can obtain a $ 14 \times 14 \times 8 $ feature mapping. For feature mapping, two different methods are used: 1) On the one hand, we pool the feature to obtain an 8-dimensional vector and provide it to the CDN module; thus, visual semantic information can be provided to align the human attributes. 2) On the other hand, the feature mapping and the outputs of HRNet are calculated as the loss of the part alignment. Supervision of the visual semantic segmentation is imposed by the pseudolabels predicted by the HRNeT \cite{Sun2018Deep} trained on the COCO Densepose dataset \cite{Rize2018DensePose}. In this way, the issues of misalignment in the CDN module can be well solved; thus, the generalizability and robustness of the learned feature can be further improved.

\subsection{Loss Function}
The person ReID task is often regarded as a person classification problem; thus, the classification loss is often calculated. In our experiments, the classification loss is also employed. To further improve the feature discrimination ability of the proposed method, a triplet loss is further added, and it is used to narrow the intraclass distance and increase the interclass loss. Finally, in the PSA module, the partial semantic alignment loss is also utilized. Thus, in total, the loss function of the MVSE method can be defined as follows:

\begin{small}
\begin{equation}
\begin{aligned}
& \mathcal{L} = \lambda_1 \mathcal{L}_{id} + \lambda_2 \mathcal{L}_{tri} + \lambda_3 \mathcal{L}_{part} ,
\end{aligned}
\end{equation}
\end{small}
where $\mathcal{L} $ is the total loss function of the MVSE method, $ \mathcal{L}_{id}$ indicates the classification loss, $\mathcal{L}_{tri}$ denotes the triplet loss, and $\mathcal{L}_{part}$ is the partial semantic alignment loss. $ \lambda_1 $, $ \lambda_2 $, and $ \lambda_3 $ are the trade-off parameters for balancing the contributions of each term. In our experiments, $ \lambda_1 $, $ \lambda_2 $, and $ \lambda_3 $ are empirically set to 0.8, 0.1, and 0.1, respectively.

In the classification loss, the merge loss is calculated by
\begin{small}
\begin{equation}
\begin{aligned}
& \mathcal{L}_{id} = \sum_{n=1}^N\{y_n \cdot \max (0, m^+ - ||C_n||)^2 \\
& + \lambda \cdot (1 - y_n) \cdot \max(0, ||C_n|| - m^-)^2\} ,
\end{aligned}
\end{equation}
\end{small}
where $N$ is the total number of training images, $ y_n $ indicates the identity to which a certain image belongs, that is, when the image belongs to identity $ n $, $ y_n = 1 $; otherwise, $ y_n = 0 $, $ n \in [1, N] $. $ \lambda = 0.5 $ is used to balance the weights between different action functions. $ m^+ $ and $ m^- $ are used to measure the length of the vector, and the set boundary values are $ m^+ = \frac{N}{N-1} $ and $ m^- = \frac{1}{N} $.

For the triple loss, an improved version can be expressed as:
\begin{small}
\begin{equation}
\begin{aligned}
& \mathcal{L}_{tri} = - \sum_{i=1}^M \sum_{a=K}^N [ \alpha + \max_{p=1...K} ||f_a^{(i)} - f_p^{(i)}||_2 \\ 
&  - \min_{\substack{n=1 \dots K\\ j=1 \dots P\\ j \neq i}}||f_a^{(i)} - f_n^{(i)}||_2 ]_+ ,
\end{aligned}
\end{equation}
\end{small}
where $N$ is the total number of training images and $M$ is the total number of person identities. $ f_a^{(i)} $, $ f_p^{(i)} $ and $ f_n^{(i)} $ are the features extracted from the anchor, positive, and negative images, respectively. $ \alpha $ is a boundary hyperparameter that is used to control the boundary values of the intraclass spacing and the interclass spacing. In our experiments, each batch has $M$ person identities, and each identity has $ K $ images. The candidate triplets are constructed from the farthest positive sample pair and the nearest negative sample pair; that is, the most difficult to distinguish positive sample pair and negative sample pair are selected.  For the partial semantic alignment loss, the popular cross-entropy loss function is used in our experiments. 

In the optimization process, the Adam optimizer \cite{Diederik2015Adam} is employed. In this way, feature discrimination and generalization can be further improved, and these features can effectively represent people with different clothes.

\section{Experiments and Discussion}
To evaluate the performance of our proposed MVSE framework, we perform experiments on four public cloth-changing person ReID datasets, namely, LTCC \cite{qian2020long}, PRCC \cite{yang2020clothing}, Celeb-reID \cite{huang2020beyond}, and NKUP \cite{wang2020benchmark}. Since the cloth-changing person ReID task is a new and challenging research topic, to the best of our knowledge, at present, there are no comprehensive experiments with any cloth-changing ReID algorithms on all four cloth-changing person ReID datasets, and this is the first work that is systematically and comprehensively assessed on these four cloth-changing person ReID datasets. The remaining section is organized as follows: 1) four public cloth-changing person ReID datasets are introduced, 2) the competitors in our experiments are described, 3) the implementation details are described, and 4) the performance evaluations and comparisons on these four public datasets are described.

\subsection{Dataset}
Few public datasets have been released in the last 2 or 3 years. In the following, four public cloth-changing person ReID datasets are introduced.

\begin{itemize} [leftmargin=*]
\item \textbf{Celeb-reID dataset (2019) \cite{huang2020beyond}}. A large-scale, long-term, person re-ID dataset, called Celeb-reID, was built in 2019. This dataset contains 34,186 images of 1,052 person IDs, where each identity contains a variety of different clothes. Moreover, more than 70\% of the images of each person contain different clothes, and the front views or profiles of people are commonly shown. It includes popular celebrities' street snap-shot images from the Internet, which were taken from a very large number of camera angles with very different backgrounds.

\item \textbf{LTCC dataset (2020) \cite{qian2020long}}. To explore the cloth-changing problem, the LTCC dataset was specifically collected with clothing changes. Moreover, different viewpoints, backgrounds, and types of severe occlusion are also included. In addition, to fit the actual situation, different people also wear the same clothes to alter their appearances, as this makes the task more challenging. In total, there are 17,119 images and 152 identities, and each ID contains 112 images on average. In terms of clothing diversity, the dataset contains 478 different outfits captured by 12 different camera views spanning more than two months.

\item \textbf{PRCC dataset (2020) \cite{yang2020clothing}}. The PRCC dataset contains three surveillance camera perspectives: A, B, and C. Among them, A and B are the same clothes in different scenes, and C contains the images collected in different clothes and at different times. In terms of the number of images, PRCC contains 33,698 pedestrian images with 221 pedestrian identities. Each identity has 152 confusing images with the same or different clothes on average.

\item \textbf{NKUP dataset (2020) \cite{wang2020benchmark}}. The NKUP dataset was collected with 15 cameras in the open environment of Nankai University, including 8 indoor cameras and 7 outdoor cameras. There are differences in viewpoint, light, resolution, body posture, season, background, and clothing across the collected images. In total, there are 107 identities and 9,738 images, and each ID contains 91 images on average.

\end{itemize}

\subsection{Competitors}
Since the cloth-changing person ReID task is a new and challenging research topic, only a few works have been published, including SPT+ASE \cite{yang2020clothing}, SE+CESD \cite{qian2020long}, and ReIDCaps \cite{huang2020beyond}. Additionally, in the cloth-changing person ReID task, traditional person ReID algorithms are often employed, such as PCB \cite{sun2018beyond}, MGN \cite{wang2018learning}, ResNet50 \cite{he2016deep}, DenseNet121 \cite{huang2017densely}, and HACNN \cite{li2018harmonious}. In the following, we introduce these methods one by one.
\begin{itemize}[leftmargin=*]

\item SPT+ASE (TPAMI 2020) \cite{yang2020clothing}. In 2020, the PRCC dataset was built, and then Yang et al. proposed an SPT+ASE module, where human contour sketching information was used to substitute for human color information. In this way, the changes in visual appearance caused by clothing changes could be reduced.

\item SE+CESD (ACCV 2020) \cite{qian2020long}. Last year, the LTCC dataset was built. To assess its difficulty level, Qian et al. proposed a shape embedding module and a clothes-eliminating shape-distillation module (SE+CESD). The former was used to encode shape information from human body keypoints, and the latter was utilized to adaptively distill the identity-relevant shape features.

\item ReIDCaps (TCSVT 2020) \cite{huang2020beyond}. In 2019, Huang et al. constructed the Celeb-reID dataset and then designed a ReIDCaps module, wherein a vector neuron concept was proposed. For each vector neuron, its direction was used to represent the changes in clothing information, and its length was utilized to identify the people.

\item ResNet50 (CVPR 2016) \cite{he2016deep}. Since the residual network has achieved satisfactory performances on many classification and retrieval tasks, it is often used in the cloth-changing person ReID task.

\item DenseNet121 (CVPR 2017) \cite{huang2017densely}. DenseNet is an upgraded version of the residual network, and it can obtain slightly better performances than those of ResNet; thus, it is often used as a competitor in the evaluation of the existing cloth-changing person ReID approaches. Moreover, it is also used as the backbone of the MVSE method.

\item HACNN (CVPR 2018) \cite{li2018harmonious}. This is a lightweight network where multiscale attention and feature representations are jointly learned.

\item PCB (ECCV 2018) \cite{sun2018beyond}. This is a part-based module, where partial human images are used to describe the person under study and provide fine-grained information for the partial person ReID task. Since it can obtain superior performances on the partial person ReID task and its local features have good robustness, this method is also used in the cloth-changing person ReID task, such as with the NKUP dataset \cite{wang2020benchmark}.

\item MGN (ACM MM 2018) \cite{wang2018learning}. Since the appearances of humans often change in the cloth-changing person ReID task, features with good robustness and generalizability are required. In the MGN, a multibranch deep network architecture is built, where one branch is built for the global feature representation and two branches are built for the local feature representation. In this way, discriminative information with various granularities can be obtained via an end-to-end feature learning strategy.
\end{itemize}

\subsection{Implementation Details}
Since the backbone of the MVSE approach is DenseNet121, DenseNet121 is used as the baseline in our experiments. In our experiments, DenseNet121 is first pretrained on the ImageNet dataset with a learning rate of $0.0001$, and the training samples of the LTCC, PRCC, Celeb-reID, and NKUP datasets are separately used to fine-tune the modules, including DenseNet121 and MVSE, whose learning rates are set to $0.001$. Note that the default settings and divisions of these datasets \cite{qian2020long, yang2020clothing, huang2020beyond, wang2020benchmark} are used to train and assess the modules. In the training procedure, the minibatch size is set to $20$, where each identity has 4 images, and the input images are resized to $ 224 \times 224$. In the optimization process, the Adam optimizer \cite{Diederik2015Adam} is employed, and $80$ epochs are required. Moreover, the learning rate is decreased by a factor of $0.1$ after $40$ and $60$ epochs.  In addition, in our experiments, $\lambda_1 = 0.8$, $\lambda_2 = 0.1$ and $\lambda_3=0.1$ in Eq. (4). Finally, the cumulative matching characteristics (CMCs), rank-1, and mean average precision (mAP) are often utilized as the evaluation metrics in person ReID tasks \cite{Gao2020DCR}; thus, we also strictly follow these metrics in our experiments.

\subsection{Performance Evaluations and Comparisons}

We first assess the performances of the MVSE method on four public cloth-changing person ReID datasets, and then we compare it with the abovementioned competitors. Among these approaches, if their codes can be obtained, ImageNet is first used to pretrain them, and then the training samples of the LTCC, PRCC, Celeb-reID, and NKUP datasets are separately used to fine-tune the modules. Finally, the testing samples of the four cloth-changing person ReID datasets are separately employed to assess their performances. If the codes are not available, the results reported by the corresponding references are used. Moreover, for a fair comparison, if the performance of a model trained by the training samples is lower than that in the corresponding reference, the results reported by the corresponding reference is also used. The results are shown in Table I. From the table, we can obtain the following observations:

\begin{table*}
\fontsize{10}{10}\selectfont
\caption{Performance evaluation and comparison on four public cloth-changing datasets, where the bold values indicate the best performance in each column.}
\renewcommand{\arraystretch}{1.5}
\begin{center}
\begin{tabular}{c|cc|cc|cc|cc}
\hline
\multirow{3}{*}{Methods} &\multicolumn{8}{c} {Datasets} \\
\cline{2-9}
& \multicolumn{2}{c|} {LTCC}& \multicolumn{2}{c|} {PRCC} & \multicolumn{2}{c|} {Celeb-reID}& \multicolumn{2}{c} {NKUP}\\
\cline{2-9}
&mAP &rank-1 &mAP &rank-1 &mAP &rank-1 &mAP &rank-1\\
\hline
ResNet-50 \cite{he2016deep} &8.4 &20.7 &- & 19.4 &5.8 &43.3 &7.9 &12.3 \\
\hline
HACNN \cite{li2018harmonious} &8.3 &20.8 &- &21.8 &9.5 &47.6 &13.7 &21.2 \\
\hline
PCB \cite{sun2018beyond} &8.8 &21.9 &- &22.9 &8.2 &37.1 &14.1 &18.7 \\
\hline
MGN \cite{wang2018learning} &10.1 &24.2 &- &25.9 &10.8 &49.0 &16.1 &20.6 \\
\hline
DenseNet-121 (Baseline) \cite{huang2017densely} &10.7 &27.2 &23.7 &18.7 &5.9 &42.9 &10.7 &15.4 \\
\hline
SE+CESD \cite{qian2020long} &11.7 &25.2 &- &- &- &- &- &- \\
\hline
SPT+ASE \cite{yang2020clothing} &- &- &- &34.4 &- &- &- &- \\
\hline
ReIDCaps \cite{huang2020beyond} &- &- &- &- &15.8 &63.0 &- &- \\
\hline
\textbf{MVSE} (ours) &\textbf{33.0} &\textbf{70.5} &\textbf{52.5} &\textbf{47.4} &\textbf{19.2} &\textbf{64.5} &\textbf{17.9} &\textbf{23.8} \\
\hline
\end{tabular}
\end{center}
\end{table*}

1) MVSE achieves the best performance regardless of the dataset and approach used, and large improvements in the mAP and rank-1 values are obtained over those of the state-of-the-art algorithms. For example, when the LTCC dataset is used, the mAP and rank-1 of the MVSE approach are 33\% and 70.5\%, respectively, but the corresponding performances of the baseline model are 10.7\% and 27.2\%, respectively, and the improvement levels can reach 21.3\% (mAP) and 43.3\% (rank-1), respectively. Similarly, the mAP and rank-1 accuracies of the MVSE approach on the PRCC dataset are 52.5\% and 47.4\%, respectively, but the corresponding mAP and rank-1 of the baseline model reach 23.7\% and 18.7\%, respectively, where the maximum improvements are 28.8\% (mAP) and 28.7\% (rank-1). Thus, MVSE can outperform DenseNet121 (baseline). The reason for this is that the visual semantic information and human attributes are embedded into DenseNet121, and then the DenseNet121 network, the CDN module, and the PSA module are jointly optimized, and the negative effect of clothing changes is reduced as much as possible. Thus, MVSE experimentally exhibits very good generalization ability, and these experimental results prove the effectiveness and robustness of the MVSE approach.

2) Among these approaches, SE+CESD, SPT+ASE, and ReIDCaps are specially designed for the cloth-changing person ReID task. In SE+CESD, a shape embedding module, as well as a clothes-eliminating shape-distillation module, is designed to eliminate unreliable clothing appearance features and focus on body shape information. In SPT+ASE, all RGB images are converted into human body contour sketches; thus, the changes in visual appearance caused by clothing changes can be reduced. In ReIDCaps, vector neuron capsules are designed, and these use the length of the vector to represent different IDs, while their orientations are used to perceive different types of clothes. However, in these approaches, low-level features are often extracted from the body shape information or human body contour sketches, but the visual high-level semantic information, which has good robustness and generalizability for the cloth-changing issue, is ignored. Moreover, the high-level human attributes and human semantic information are jointly used in the MVSE method. Thus, the performance of MVSE is much better than those of SE+CESD, SPT+ASE, and ReIDCaps. For example, on the LTCC dataset, the mAP and rank-1 accuracies of the MVSE method are 33\% and 70.5\%, respectively, but the mAP and rank-1 accuracies of SE+CESD are 11.7\% and 25.2\%, respectively; thus, the improvements can reach 22.3\% (mAP) and 45.3\% (rank-1). Similarly, the mAP and rank-1 accuracies of the MVSE method on the PRCC dataset can reach 52.5\% and 47.4\%, respectively, and the rank-1 accuracy of SPT+ASE is 34.4\%; thus, the improvement is 13\% (rank-1). Finally, when the Celeb-reID dataset is used, the mAP and rank-1 accuracies of the MVSE approach are 19.2\% and 64.5\%, respectively, but the mAP and rank-1 accuracies of ReIDCaps can reach 15.8\% and 63\%, respectively, where the improvements are 3.4\% (mAP) and 1.5\% (rank-1). Thus, these experimental results can further prove the effectiveness and robustness of MVSE.

3) For ResNet50, DenseNet121, PCB, MGN, and HACNN, the top two modules are widely used in many machine learning tasks, but they are also often assessed on the cloth-changing person ReID task. Others are specially designed for a person ReID task where a discrimination feature can be learned. Although these modules can achieve good performances in many related tasks, when they are directly applied to the cloth-changing person ReID task, their performances are unsatisfactory, and they are much worse than that of MVSE. For example, when the PRCC dataset is used, the rank-1 accuracies of ResNet50, HACNN, PCB, MGN, DenseNet121, and MVSE are 19.4\%, 21.8\%, 22.9\%, 25.9\%, 18.7\%, and 47.4\%, respectively, and the corresponding improvements achieved by the MVSE method are 28\%, 25.6\%, 24.5\%, 21.5\%, and 28.7\%, respectively. Similarly, the mAP accuracies of ResNet50, HACNN, PCB, MGN, DenseNet121, and MVSE on the Celeb-reID dataset are 5.8\%, 9.5\%, 8.2\%, 10.8\%, 5.9\%, and 19.2\%, respectively, and the corresponding improvements achieved by the MVSE method are 13.4\%, 9.7\%, 11\%, 8.4\%, and 13.3\%, respectively. In addition, we also find that the performances of SE+CESD, SPT+ASE, and ReIDCaps are better than those of ResNet50, DenseNet121, PCB, MGN, and HACNN; for example, the rank-1 accuracies of SPT+ASE and MGN (the best performances of the features not designed for the cloth-changing person ReID task) are 34.4\% and 25.9\%, respectively, and the former method outperforms the latter. We can draw similar conclusions on other datasets.

4) If more camera views are used to capture the people in the gallery dataset, then generally speaking, when we query a person, we have more chances to obtain the correct rank-1 results. Therefore, in the LTCC dataset, 12 different camera views are utilized, and in the Celeb-reID dataset, a very large number of camera angles with very different backgrounds are employed; thus, the rank-1 accuracies of MVSE on the LTCC and Celeb-reID datasets are 70.5\% and 64.5\%, respectively, and these are the top two results among all four datasets. In the PRCC dataset, only two cameras are used, but the changes in backgrounds, viewpoints, and lighting are not large; thus, the rank-1 accuracy on this dataset is relatively good (47.4\%). In the NKUP dataset, 8 indoor cameras and 7 outdoor cameras are utilized, but there are large differences in viewpoint, light, resolution, body posture, season, background, and clothing for the collected images; thus, the rank-1 accuracy of MVSE is worst on this dataset (23.8\%). In addition, the mAP accuracies on these datasets are also very different, and this evaluation metric can reflect the level of difficulty of each cloth-changing person ReID dataset. For example, in the PRCC dataset, only two cameras are utilized; moreover, the changes in backgrounds, viewpoints, and lighting are small; thus, the mAP of MVSE is best on this dataset (52.5\%). Similarly, in the NKUP dataset, 15 indoor and outdoor cameras are employed, and the differences between different images are large; thus, it is the most difficult dataset, and the mAP of MVSE is only 17.9\%.

\section{Ablation Study}
An ablation study is performed using the MVSE model to analyze the contribution of each component. In this investigation, five aspects are considered: 1) the effectiveness of the MGR module, 2) the advantages of the CDN module, 3) the benefits of the PSA module, 4) a convergence analysis, and 5) a qualitative visualization. In the following, we discuss these five aspects separately.

\subsection{Effectiveness of the MGR Module}
In this section, we demonstrate the effectiveness of the MGR module. Since DenseNet121 is used as the backbone of MVSE, in our experiments, DenseNet121 is used as the baseline. Among the few existing cloth-changing person ReID methods, the problem of changes in the characteristics of clothes is often solved by global features, but local features, including information such as the head, feet, bare skin, and carried objects, can often provide effective discrimination ability. Thus, the multigranular learning strategy is used in MVSE, where the feature map is divided into different feature blocks, and then we can obtain the features $F_1$, $F_2$, $F_3$, $F_4$, $F_5$, and $F_6$ (note that these features are learned by the baseline module). Finally, these features are concatenated to obtain the MGR features. Their results are shown in Table II, where the MGR indicates that five individual features are concatenated to form the fusion feature, and from it, we can obtain the following observations:

\begin{table*}
\fontsize{10}{10}\selectfont
\caption{Effectiveness of the MGR module, where four public cloth-changing datasets are employed, and the bold values indicate the best performance in each column.}
\renewcommand{\arraystretch}{1.4}
\begin{center}
\begin{tabular}{c|cc|cc|cc|cc}
\hline
\multirow{3}{*}{Methods} &\multicolumn{8}{c} {Datasets} \\
\cline{2-9}
& \multicolumn{2}{c|} {LTCC}& \multicolumn{2}{c|} {PRCC} & \multicolumn{2}{c|} {Celeb-reID}& \multicolumn{2}{c} {NKUP}\\
\cline{2-9}
&mAP &rank-1 &mAP &rank-1 &mAP &rank-1 &mAP &rank-1\\
\hline
$ F_1 $ &10.7 &27.2 &23.1 & 18.5 &2.9 &29.4 &3.5 &6.6 \\
$ F_2 $ &6.5 &15.6 &14.5 &10.1 &2.9 &29.1 &3.0 &4.2 \\
$ F_3 $ &8.3 &24.9 &35.2 &29.4 &2.1 &22.5 &3.2 &5.7 \\
$ F_4 $ &5.7 &7.6 &10.0 &5.4 &2.7 &26.7 &3.8 &6.1 \\
$ F_5 $ &6.8 &24.2 &15.7 &11.4 &2.0 &23.1 &2.9 &3.3 \\
$ F_6 $ &8.7 &21.9 &34.8 &29.6 &1.8 &17.6 &4.5 &5.5 \\
\hline
\textbf{MGR} &\textbf{14.0} &\textbf{37.5} &\textbf{38.4} &\textbf{33.0} &\textbf{5.9} &\textbf{38.8} &\textbf{5.0} &\textbf{8.5} \\
\hline
\end{tabular}
\end{center}
\end{table*}

1) The MGR is very effective and robust, and it can obtain the best performance regardless of the dataset and individual features used. Furthermore, large improvements in mAP and rank-1 are obtained over those yielded by different individual features. For example, on the LTCC dataset, the best rank-1 accuracy is 27.2\% among all features, but the best rank-1 accuracy obtained by the MVSE method is 37.5\%, where the improvement is 10.3\%. Similarly, on the Celeb-reID dataset, the $F_1$ feature can obtain the best rank-1 accuracy (29.4\%), and the rank-1 accuracy of the MVSE method can reach 38.8\%, where the improvement is 9.4\%. We can draw similar conclusions from the other datasets. These experimental results show that the MGR is very effective in describing cloth-changing people, and robust features can be obtained.

2) When individual features $ F_1 $ to $F_6$ are used to represent the cloth-changing person, their performances are not robust and stable when different datasets are utilized. For example, on the LTCC and Celeb-reID datasets, the global feature ($F_1$) can obtain the best mAP and rank-1 accuracies among all individual features, but on the PRCC dataset, local feature $F_3$ can obtain the best mAP among all individual features; then, on the NKUP dataset, local feature $F_6$ achieves the best mAP among all individual features. Thus, these experimental results show that both the global feature and the local features are very useful for describing cloth-changing persons, but they are not stable and robust. When these features are individually used to represent a cloth-changing person, their performances are unsatisfactory, but these features are complementary. When the MGR module is employed, the performance is greatly improved regardless of the dataset utilized.

\subsection{Advantages of the CDN Module}

In this section, we assess the effectiveness of the CDN module on four public cloth-changing person ReID datasets. Although the global feature and the local features are fully used to represent a cloth-changing person in the MGR module, the differences between the appearances of humans in the cloth-changing person ReID task are very large, and the generalized learning ability with respect to clothing changes needs to be improved; thus, the CDN module is designed to improve the feature discrimination ability and the clothing desensitization ability of the proposed method. To fully assess the advantages of the CND module, all feature mappings $F_1$, $F_2$, $F_3$, $F_4$, $F_5$, and $F_6$ are further fed into the CDN module, and then we can obtain new discriminative features $F_1+CDN$, $F_2+CDN$, $F_3+CDN$, $F_4+CDN$, $F_5+CDN$, and $F_6+CDN$, respectively. Their results are given in Fig. 3 and Fig. 4. Note that when the MGR module and the CDN module are employed, the results are shown by the labels ``MGR`` and ``MGR+CDN``. From these figures, we can observe that:

1) The CDN module is very effective and robust. When the CDN module is used for the individual features, the mAP and rank-1 accuracies can be obviously improved regardless of the dataset and individual features used. For example, when the $F_1$ feature is used, the rank-1 (mAP) accuracies on the LTCC, PRCC, Celeb-reID and NKUP datasets are 27.2\% (10.7\%), 18.5\% (23.1\%), 29.4\% (2.9\%), and 6.6\% (3.5\%), respectively, but the corresponding rank-1 (mAP) accuracies of $F_1+CDN$ reach 60.5\% (23.5\%), 26.5\% (32.6\%), 50.9\% (9.6\%) and 19.0\% (11.5\%), respectively, where the improvements are 33.3\% (12.8\%), 8.0\% (9.5\%), 21.5\% (6.7\%), and 12.4\% (8.0\%), respectively. Similarly, when the $F_4$ feature is used, the rank-1 (mAP) accuracies on the LTCC, PRCC, Celeb-reID and NKUP datasets are 7.6\% (5.7\%), 5.4\% (10.0\%), 26.7\% (2.7\%), and 6.1\% (3.8\%), respectively, but the corresponding rank-1 (mAP) accuracies of $F_4+CDN$ reach 38.3\% (14.3\%), 25.0\% (29.5\%), 53.2\% (11.5\%), and 13.0\% (8.0\%), respectively, where the improvements are 30.7\% (8.6\%), 19.6\% (19.5\%), 26.5\% (8.8\%), and 6.9\% (4.2\%), respectively. The reason why the CDN module is very effective and robust is that the human attributes are high-level semantic features that are insensitive to clothing changes; thus, the extracted features can have good generalization capabilities with regard to cloth-changing persons.

2) When the MGR module is used, the performance is improved over those of the individual features, such as $F_1$, $F_2$, $F_3$, $F_4$, $F_5$, and $F_6$, but sometimes, the improvement is limited. For example, on the NKUP dataset, the mAP and rank-1 accuracies of the MGR are 5.0\% and 8.5\%, and the best mAP and rank-1 accuracies of the individual features are 4.5\% and 6.6\%, respectively; thus, the improvements are only 0.5\% (mAP) and 1.9\% (rank-1). However, when the CDN module is further utilized, the mAP and rank-1 accuracies of MGR+CDN are 15.1\% and 21.5\%, respectively, and the mAP and rank-1 accuracies of the MGR+the individual features are 12.1\% and 19.0\%, respectively; thus, the improvements are only 3.0\% (mAP) and 2.5\% (rank-1). Therefore, these experimental results can further prove the effectiveness of the CDN module.

\begin{figure}[h]
\begin{minipage}{0.49\linewidth}
\centerline{\includegraphics[width=1.6in,height = 1.3in]{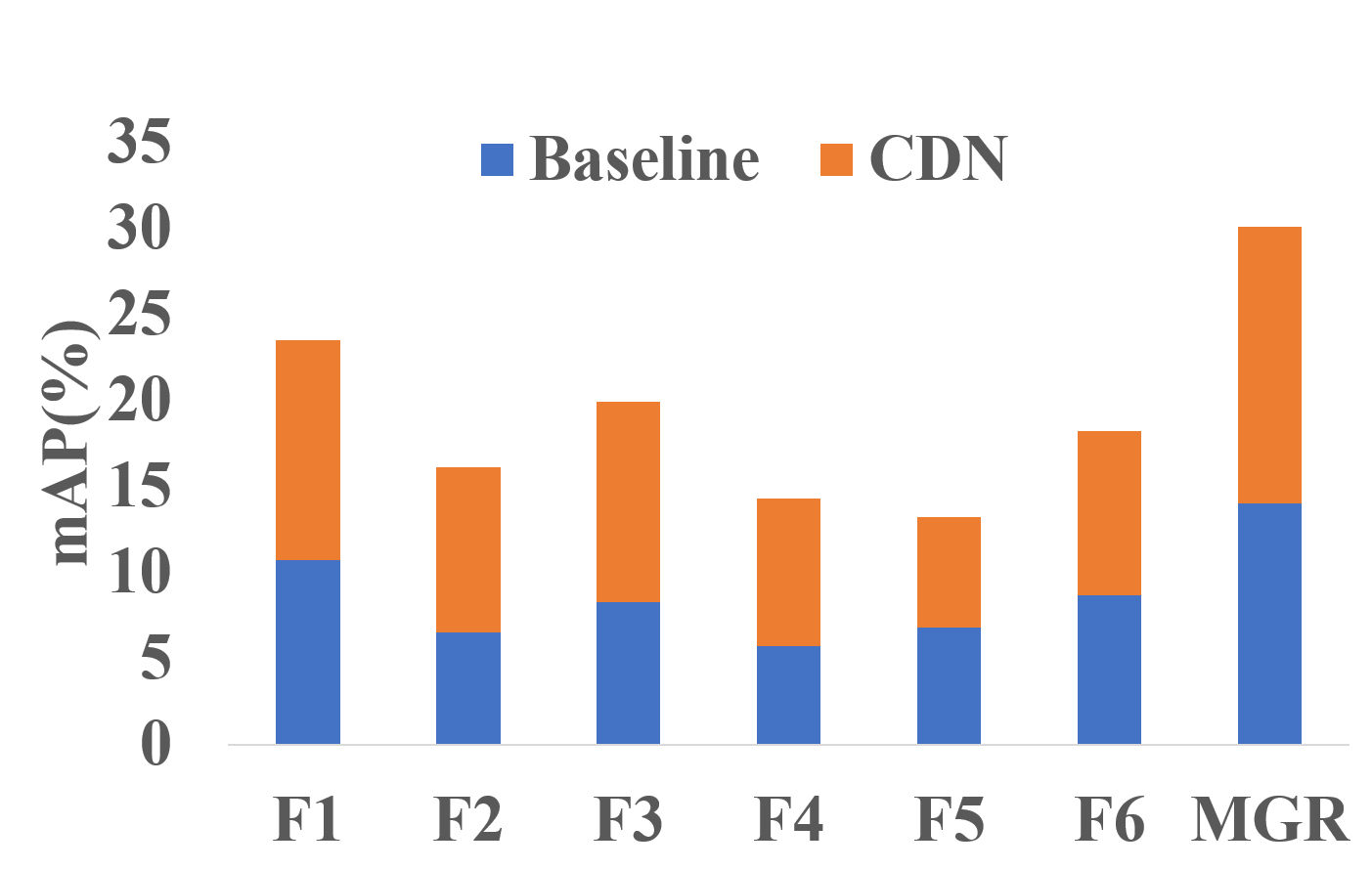}}
\centerline{(a) LTCC}
\end{minipage}
\hfill
\begin{minipage}{0.49\linewidth}
\centerline{\includegraphics[width=1.6in,height = 1.3in]{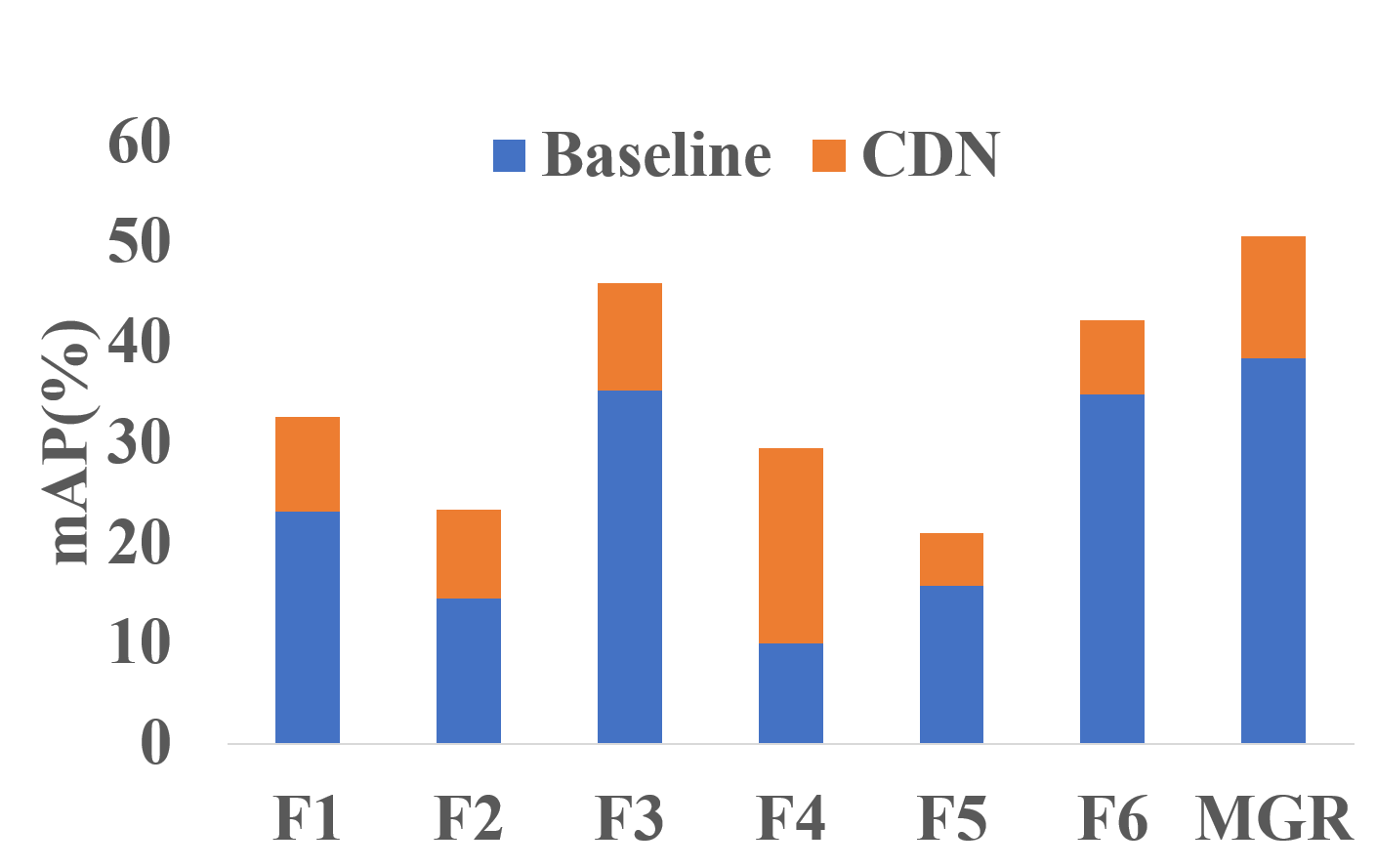}}
\centerline{(b) PRCC}
\end{minipage}
\\
\begin{minipage}{0.49\linewidth}
\centerline{\includegraphics[width=1.6in,height = 1.3in]{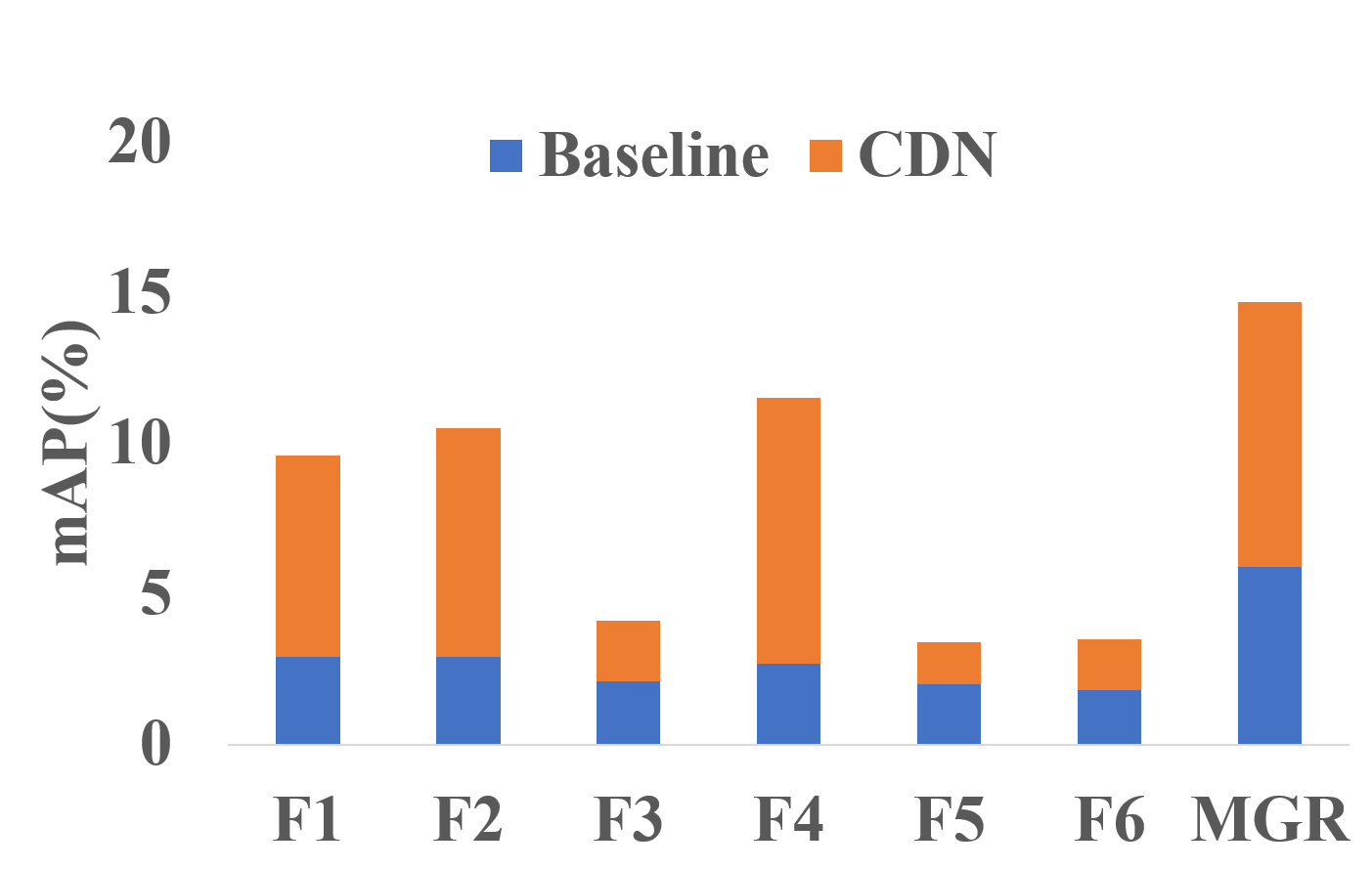}}
\centerline{(c) Celeb-reID}
\end{minipage}
\hfill
\begin{minipage}{0.49\linewidth}
\centerline{\includegraphics[width=1.6in,height = 1.3in]{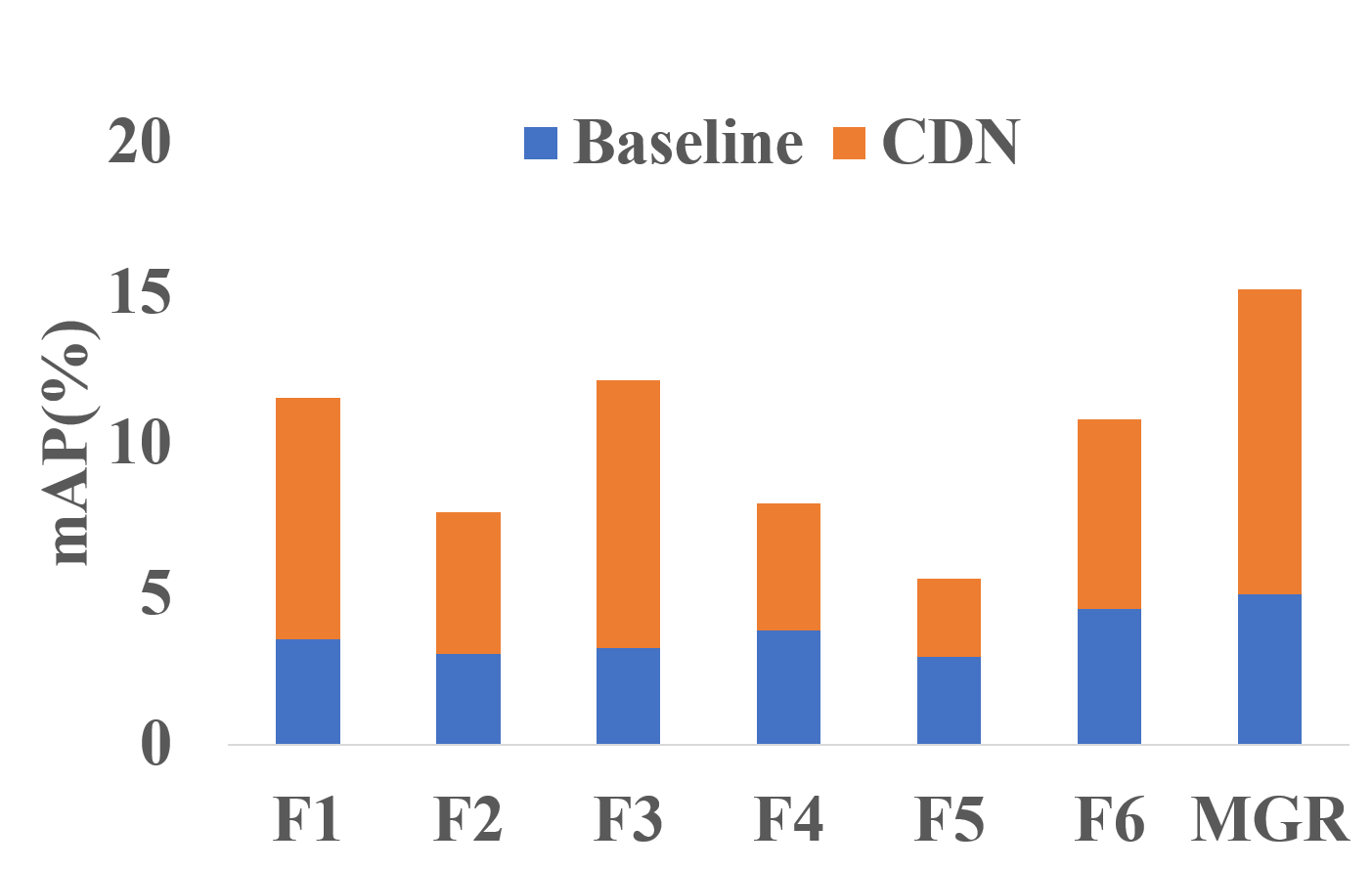}}
\centerline{(d) NKUP}
\end{minipage}
\caption{Advantages of the CDN module, where four public cloth-changing datasets are utilized, and the mAP is used as the evaluation metric. Note that the blue bar indicates the results of the baseline, and the yellow bar denotes the improvements over the baseline when the CDN module is further used.}
\end{figure}

\begin{figure}[h]
\begin{minipage}{0.49\linewidth}
\centerline{\includegraphics[width=1.6in,height = 1.3in]{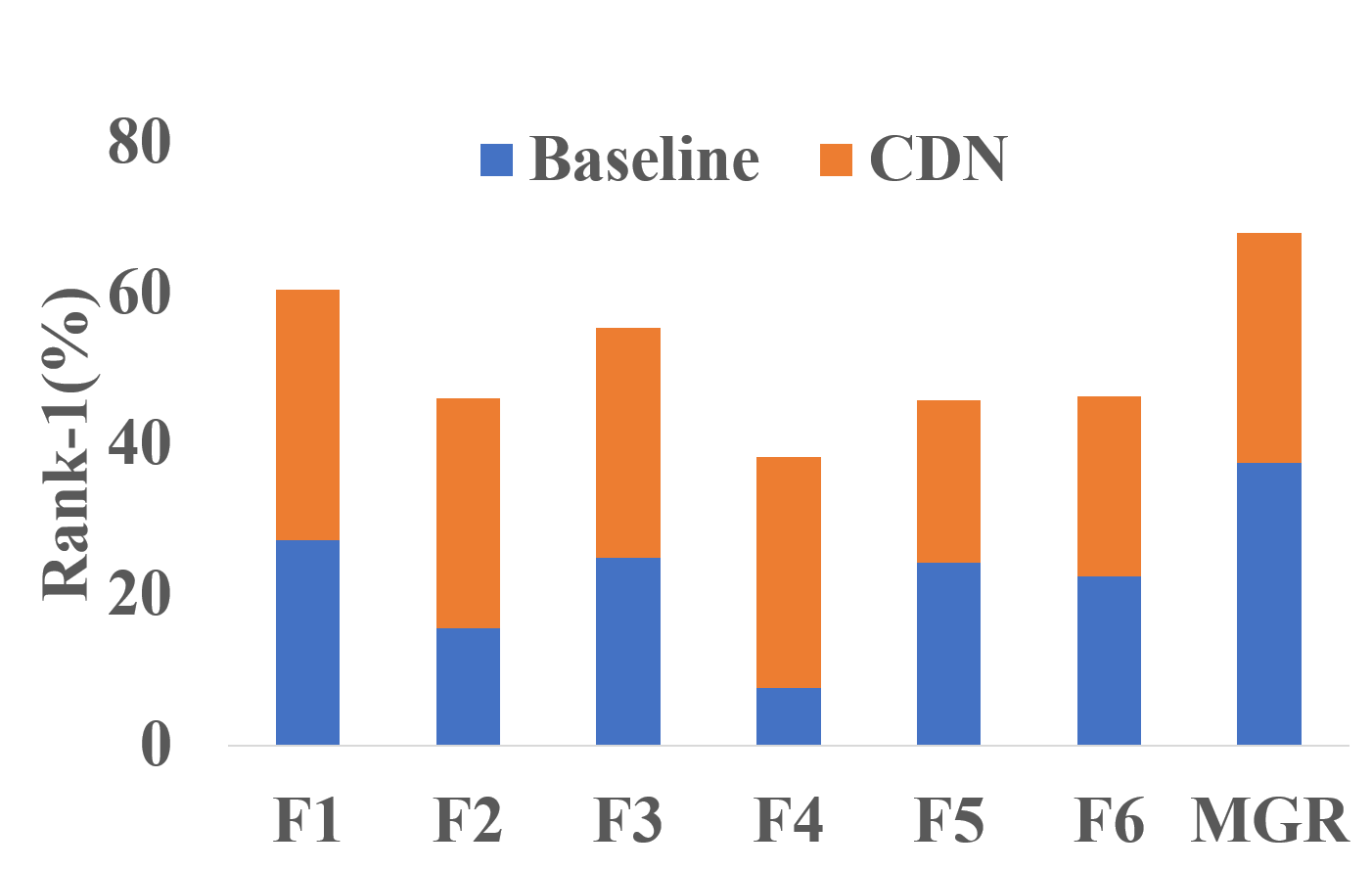}}
\centerline{(a) LTCC}
\end{minipage}
\hfill
\begin{minipage}{0.49\linewidth}
\centerline{\includegraphics[width=1.6in,height = 1.3in]{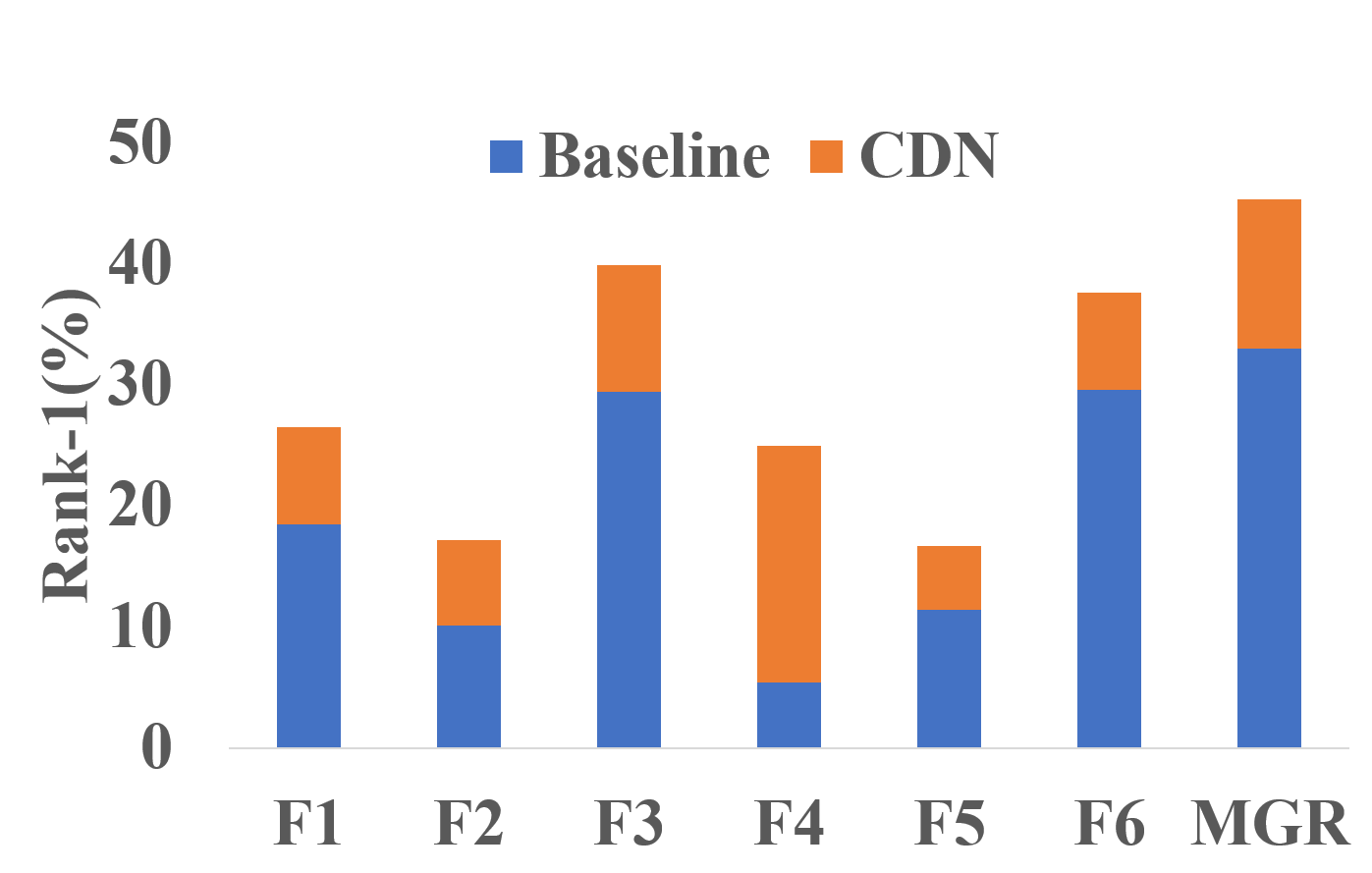}}
\centerline{(b) PRCC}
\end{minipage}
\\
\begin{minipage}{0.49\linewidth}
\centerline{\includegraphics[width=1.6in,height = 1.3in]{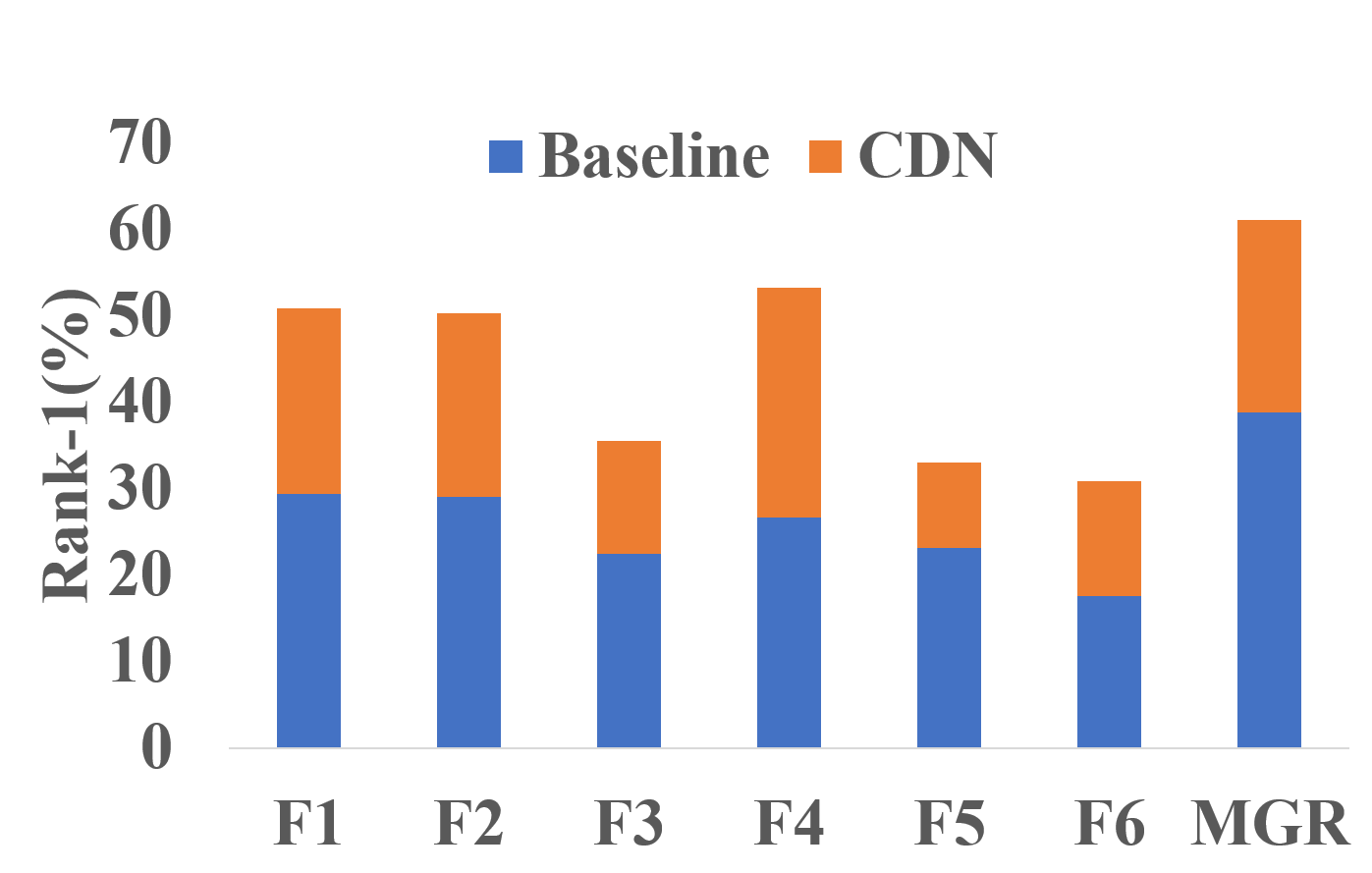}}
\centerline{(c) Celeb-reID}
\end{minipage}
\hfill
\begin{minipage}{0.49\linewidth}
\centerline{\includegraphics[width=1.6in,height = 1.3in]{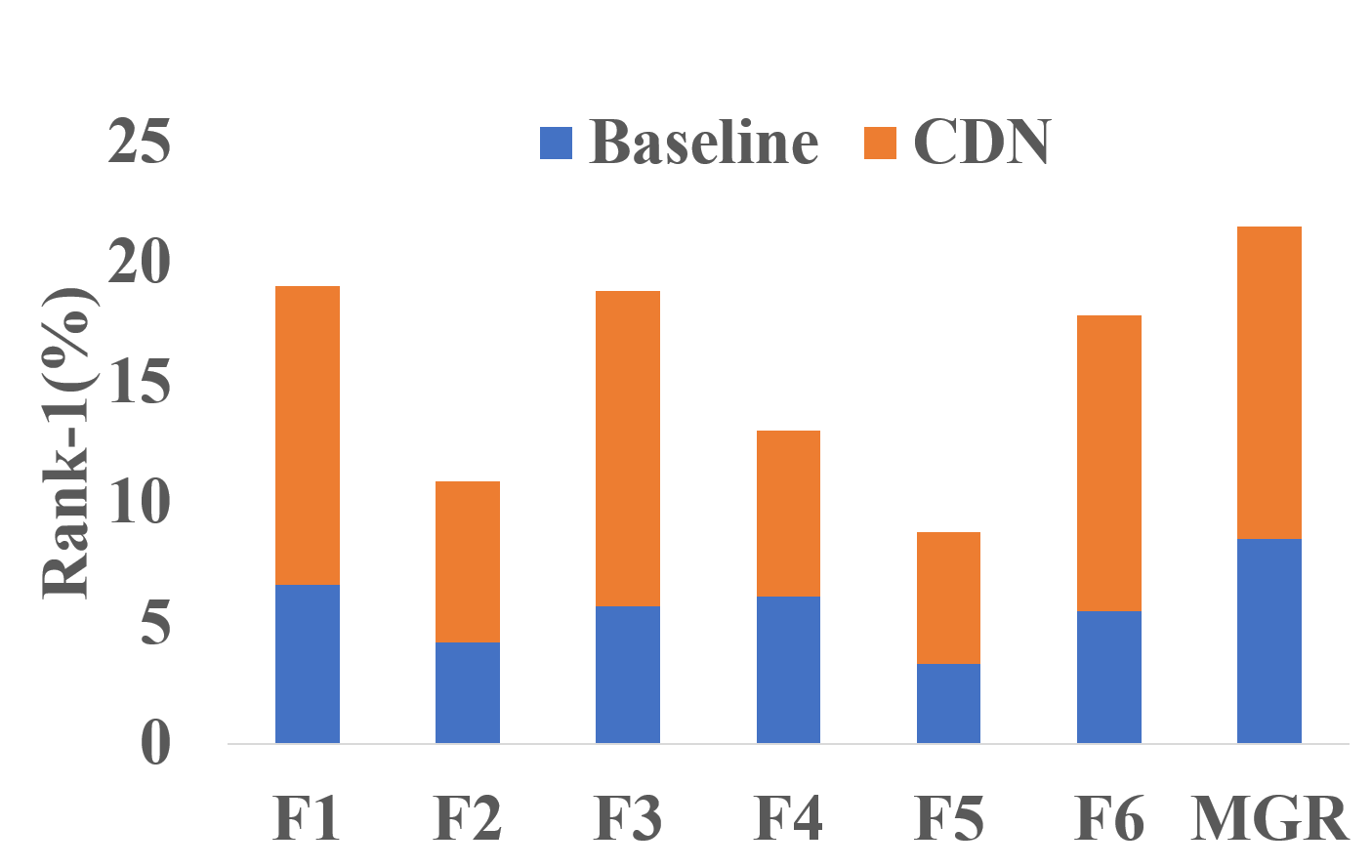}}
\centerline{(d) NKUP}
\end{minipage}
\caption{Advantages of the CDN, module where four public cloth-changing datasets are utilized, and rank-1 is used as the evaluation metric. Note that the blue bar indicates the results of the baseline, and the yellow bar denotes the improvements over the baseline when the CDN module is further used.}
\end{figure}

\subsection{Benefits of the PSA Module}

To validate the effectiveness of the PSA module, we perform experiments on the four public cloth-changing person ReID datasets, and their results are given in Table III and Fig. 5. Note that when the global feature ($F_1$) is fed into DenseNet121, the module is considered the baseline. When all individual features are combined by the MGR module, it is denoted as MGR. Moreover, when the CDN module is further utilized, it is called MGR+CDN. Finally, based on MGR+CDN, the PSA module is further employed, and this method is called MGR+CDN+PSA. From the results, we can see that:

1) When the PSA module is used, the performance of MGR+CDN+PSA is greatly improved over that of MGR+CDN. For example, when the Celeb-reID dataset is used, the mAP and rank-1 accuracies of MGR+CDN+PSA are 19.2\% and 64.5\%, respectively, and the mAP and rank-1 accuracies of MGR+CDN are 14.7\% and 61.1\%, respectively, where the improvements 4.5\% (mAP) and 3.4\% (rank-1). We can draw similar conclusions from the other datasets. In addition, when the CMC curves of rank-5 and rank-10 in Fig. 5 are used as the evaluation metrics, we can obtain the same results. The reason why the PSA module can be successful is that changes in the backgrounds, viewpoints, poses, observation perspectives and lighting of images often occur, but human attributes alone cannot fully describe the appearance of a human, and these attributes are often not aligned when the pose or observation perspective changes. However, in the PSA module, human visual-semantic information can be obtained, which is then used to align the human attributes; thus, the extracted feature is robust and effective and can effectively solve the problem of changes in pose or observation perspective.

2) In the MGR+CDN+PSA module, multigranular features, high-level human attributes, and human semantic information are jointly learned in a unified framework; thus, these modules are complementary, and they can promote each other. When the MGR module, the CDN module, and the PSA module are embedded into the baseline step by step, their combined performance can yield a stable improvement. In particular, when we compare MGR+CDN+PSA with the baseline, the improvements are very obvious. For example, on the LTCC dataset, the mAP and rank-1 accuracies of MGR+CDN+PSA are 33\% and 70.5\%, respectively, but the mAP and rank-1 accuracies of the baseline are 10.7\% and 27.2\%, respectively, and the improvements are 22.3\% (mAP) and 43.3\% (rank-1). Similarly, on the PRCC dataset, the corresponding improvements are 29.4\% (mAP) and 28.9\% (rank-1).

\begin{table*}
\fontsize{10}{10}\selectfont
\caption{Benefits of the PSA module ,where four public cloth-changing datasets are utilized. Notice that the DenseNet121 classification module is used as the baseline, and our proposed MGR, CDN and PSA modules are embedded into the baseline step by step.}
\renewcommand{\arraystretch}{1.7}
\begin{center}
\begin{tabular}{c|cc|cc|cc|cc}
\hline
\multirow{3}{*}{Methods} &\multicolumn{8}{c} {Datasets} \\
\cline{2-9}
& \multicolumn{2}{c|} {LTCC} & \multicolumn{2}{c|} {PRCC} & \multicolumn{2}{c|} {Celeb-reID} & \multicolumn{2}{c} {NKUP}\\
\cline{2-9}
&mAP &rank-1 &mAP &rank-1 &mAP &rank-1 &mAP &rank-1\\
\hline
DenseNet121 (F1) (Baseline) &10.7 &27.2 &23.1 &18.5 &2.9 &29.4 &3.5 &6.6\\

MGR &14.0 &37.5 &35.2 &33.0 &5.9 &38.8 &5.0 &8.5\\
MGR+CDN &30.1 &68.0 &50.6 &45.3 &14.7 &61.1 &15.1 &21.5\\
MGR+CDN+PSA (MVSE) &33.0 &70.5 &52.5 &47.4 &19.2 &64.5 &17.9 &23.8 \\
\hline
\end{tabular}
\end{center}
\end{table*}

\begin{figure}[h]
\begin{minipage}{0.49\linewidth}
\centerline{\includegraphics[width=1.8in,height = 1.4in]{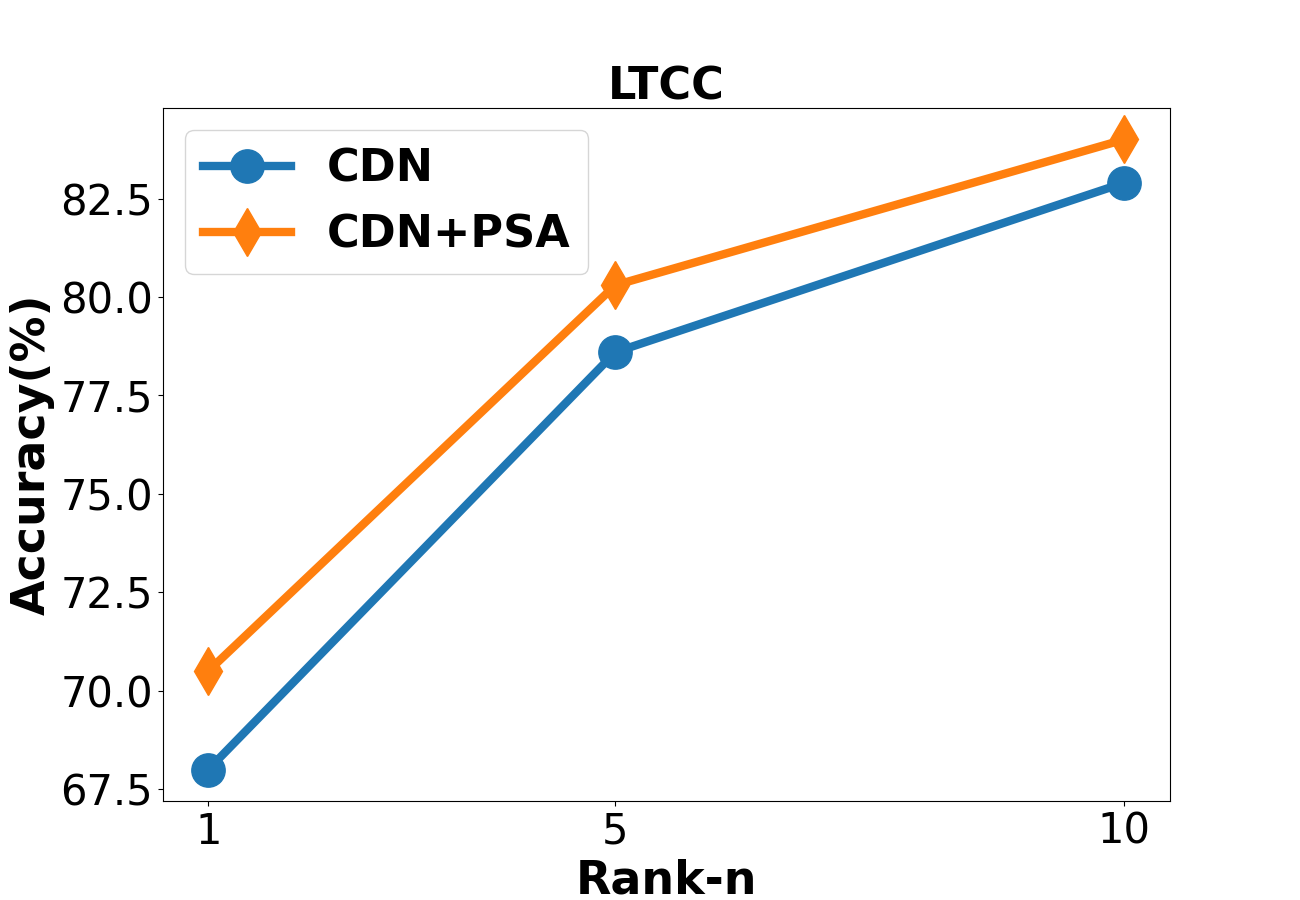}}
\end{minipage}
\hfill
\begin{minipage}{0.49\linewidth}
\centerline{\includegraphics[width=1.8in,height = 1.4in]{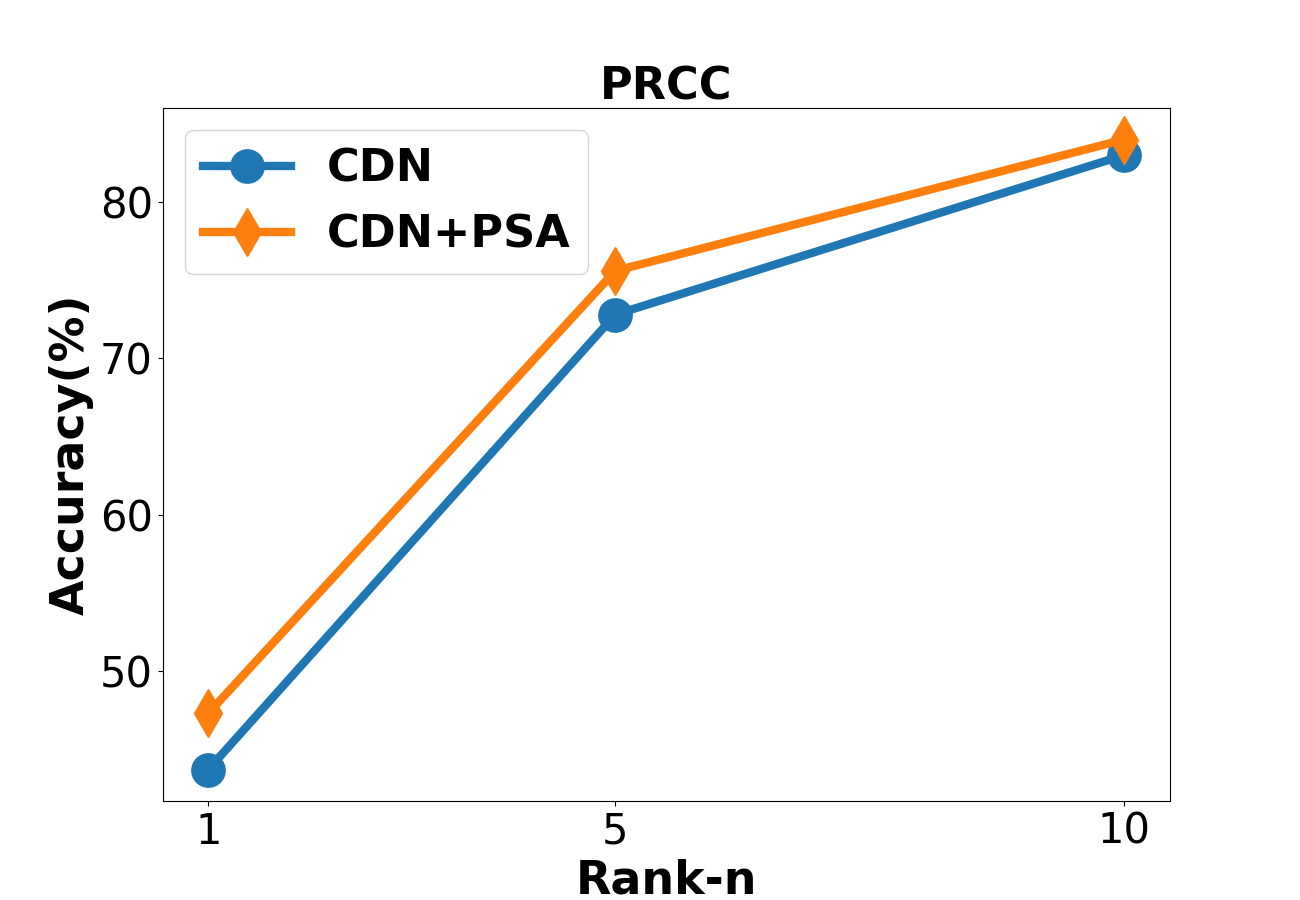}}
\end{minipage}
\\
\begin{minipage}{0.49\linewidth}
\centerline{\includegraphics[width=1.8in,height = 1.4in]{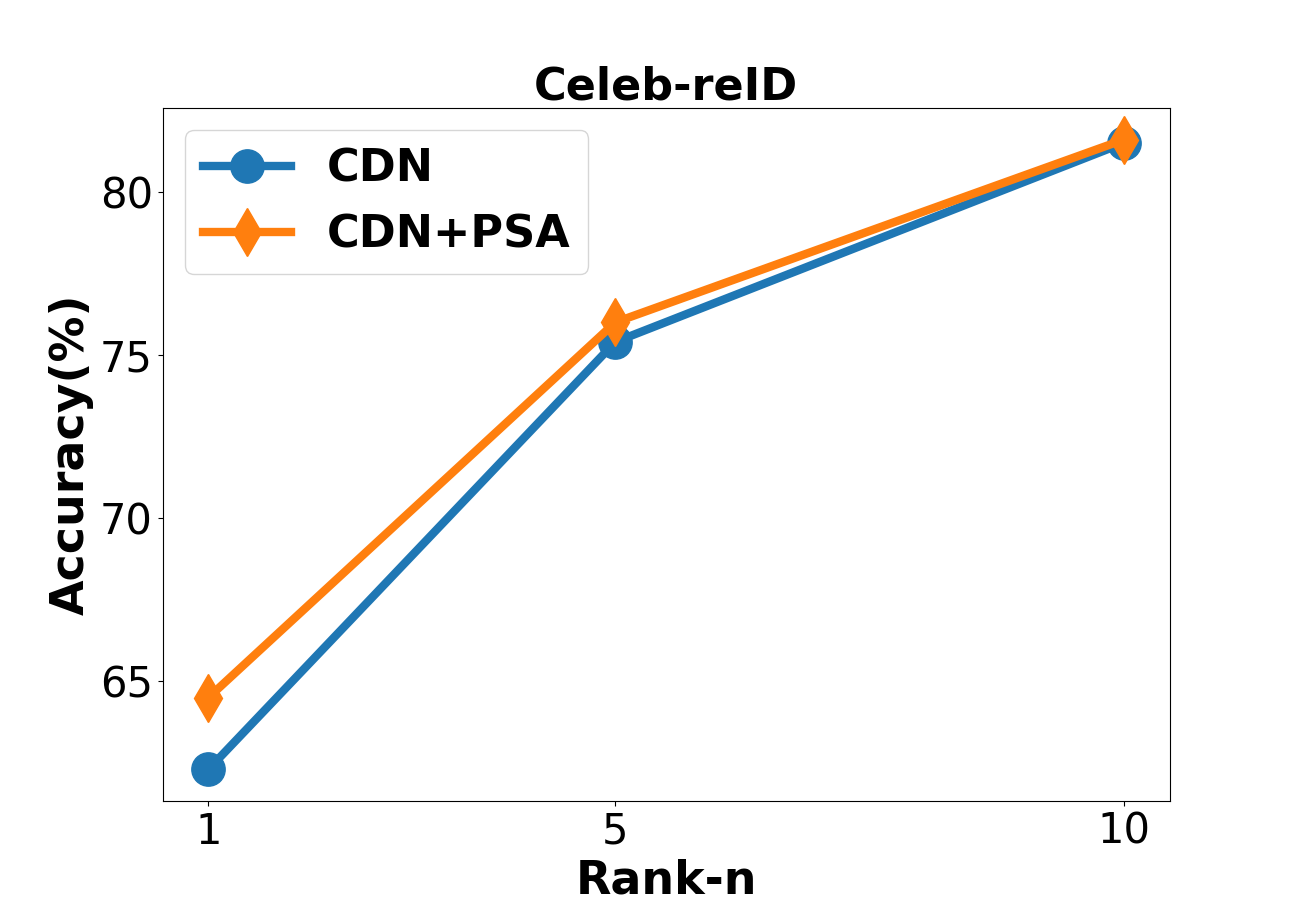}}
\end{minipage}
\hfill
\begin{minipage}{0.49\linewidth}
\centerline{\includegraphics[width=1.8in,height = 1.4in]{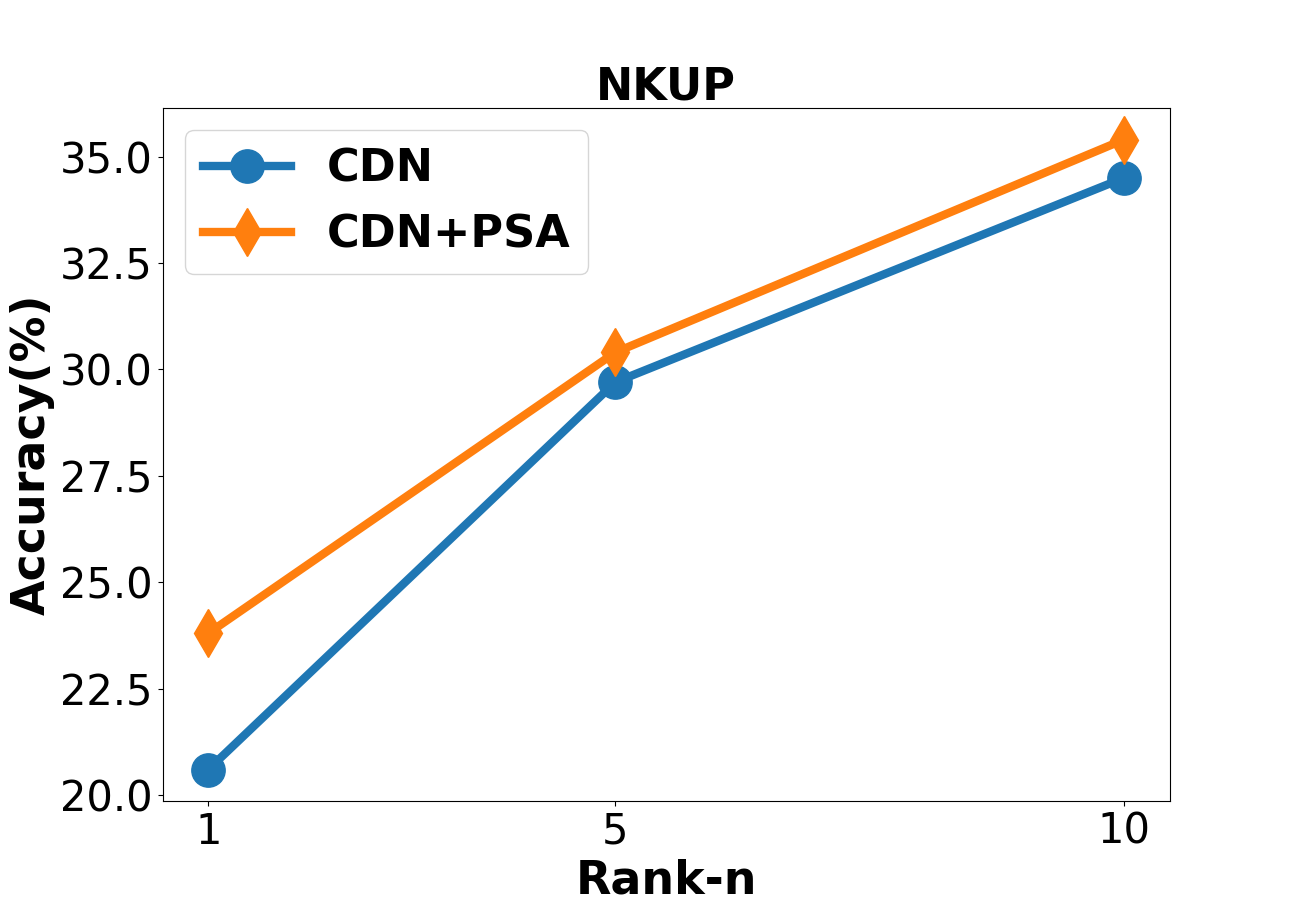}}
\end{minipage}
\caption{Advantage analysis of the PSA module by using CMC curves for the LTCC, PRCC, Celeb-reID and NKUP datasets. }
\end{figure}

\subsection{Convergence Analysis}
In this section, we evaluate the convergence of the proposed MVSE method on four public cloth-changing person ReID datasets, including LTCC, PRCC, Celeb-reID, and NKUP, and their convergence curves are shown in Fig. 6. Note that in the MVSE framework, the MGR module is replaced by $F_1$, $F_2$, $F_3$, $F_4$, $F_5$, and $F_6$. From these figures, the convergence speeds of the MVSE method are very fast when different individual features are used. In the optimization process, only 40-50 epochs are required for all datasets, and the convergence curves can be stable regardless of the dataset utilized. Thus, this can further prove the effectiveness of the MVSE method.

\begin{figure}[h]
\begin{minipage}{0.49\linewidth}
\centerline{\includegraphics[width=1.8in,height = 1.4in]{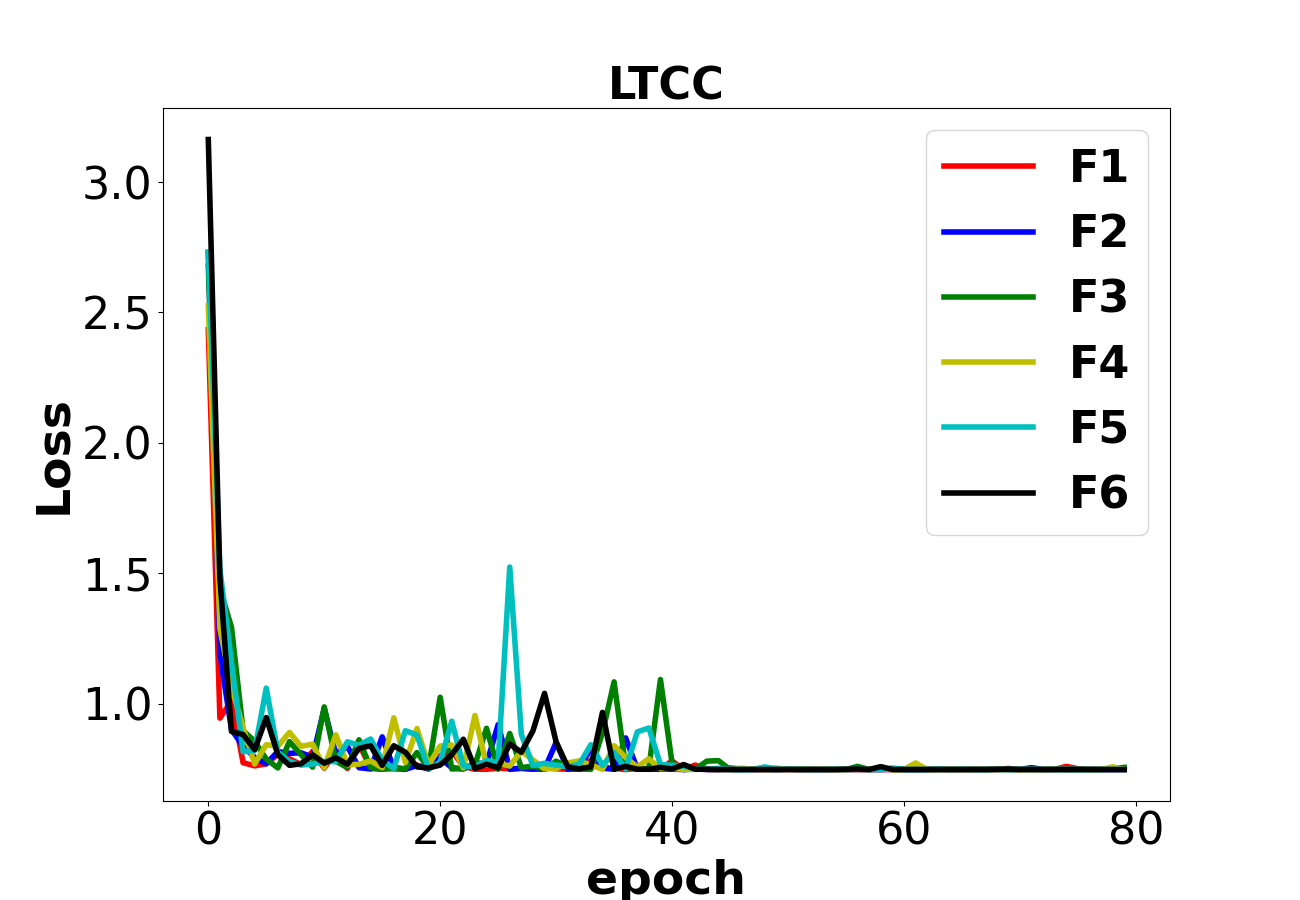}}
\end{minipage}
\hfill
\begin{minipage}{0.49\linewidth}
\centerline{\includegraphics[width=1.8in,height = 1.4in]{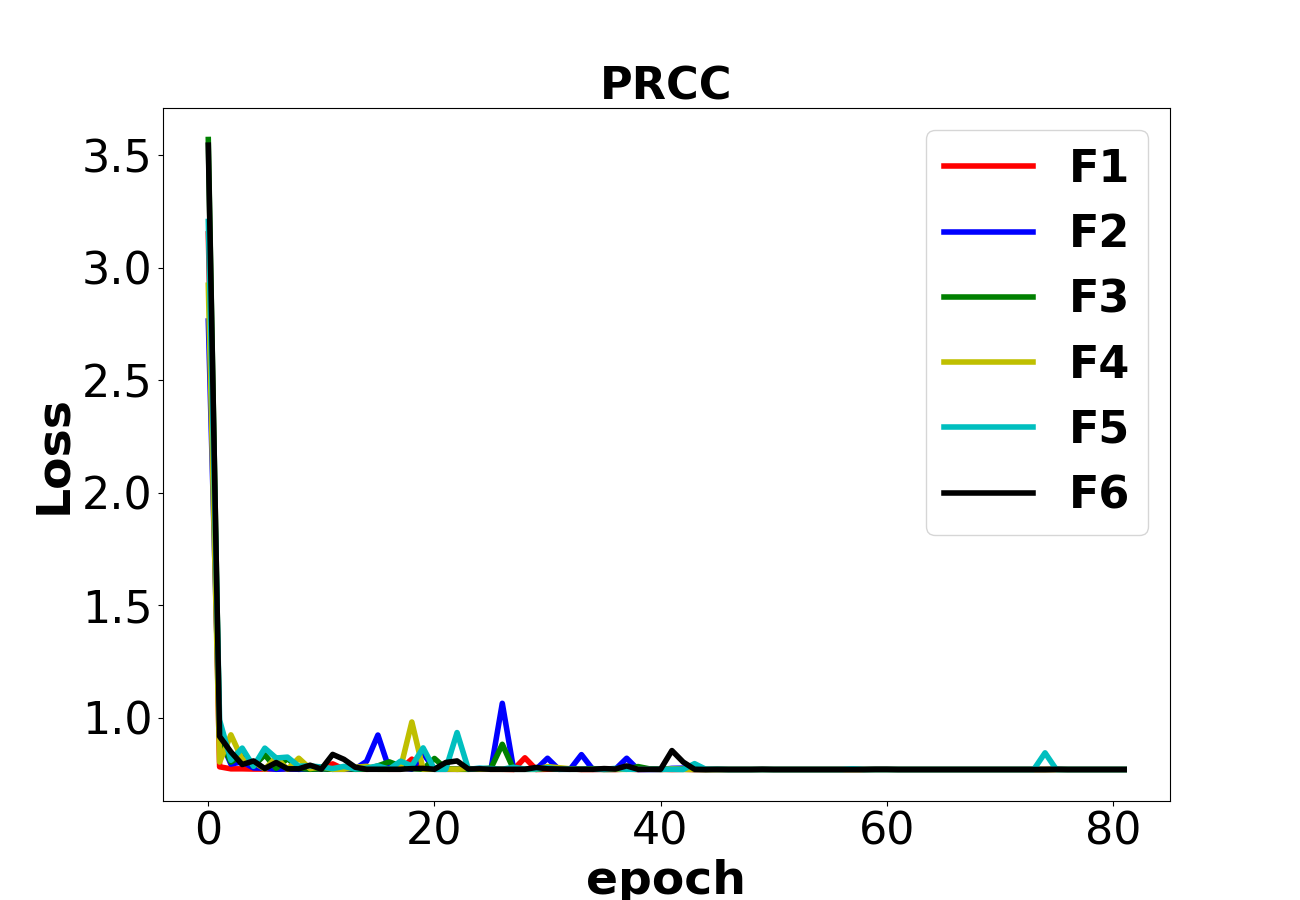}}
\end{minipage}
\\
\begin{minipage}{0.49\linewidth}
\centerline{\includegraphics[width=1.8in,height = 1.4in]{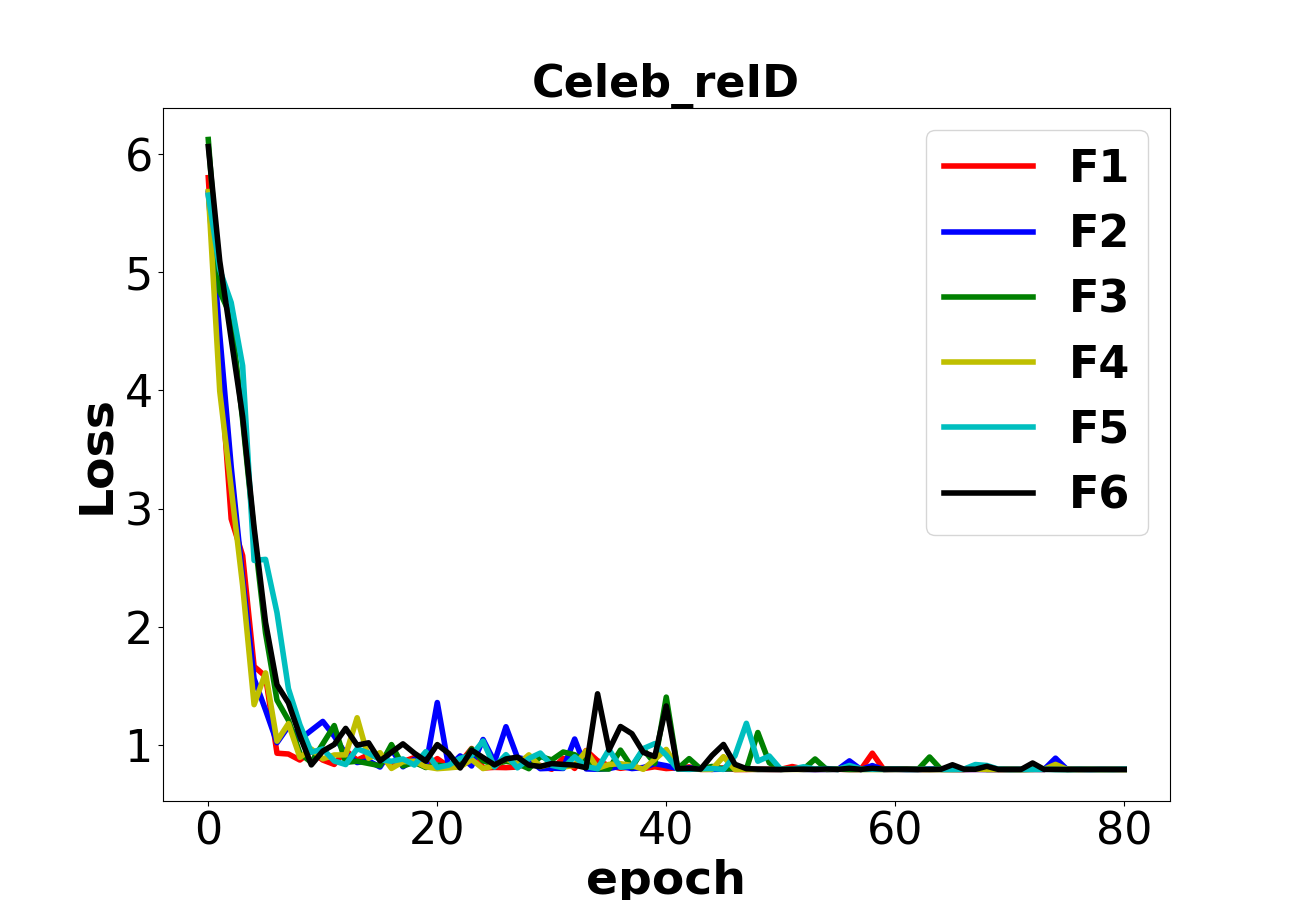}}
\end{minipage}
\hfill
\begin{minipage}{0.49\linewidth}
\centerline{\includegraphics[width=1.8in,height = 1.4in]{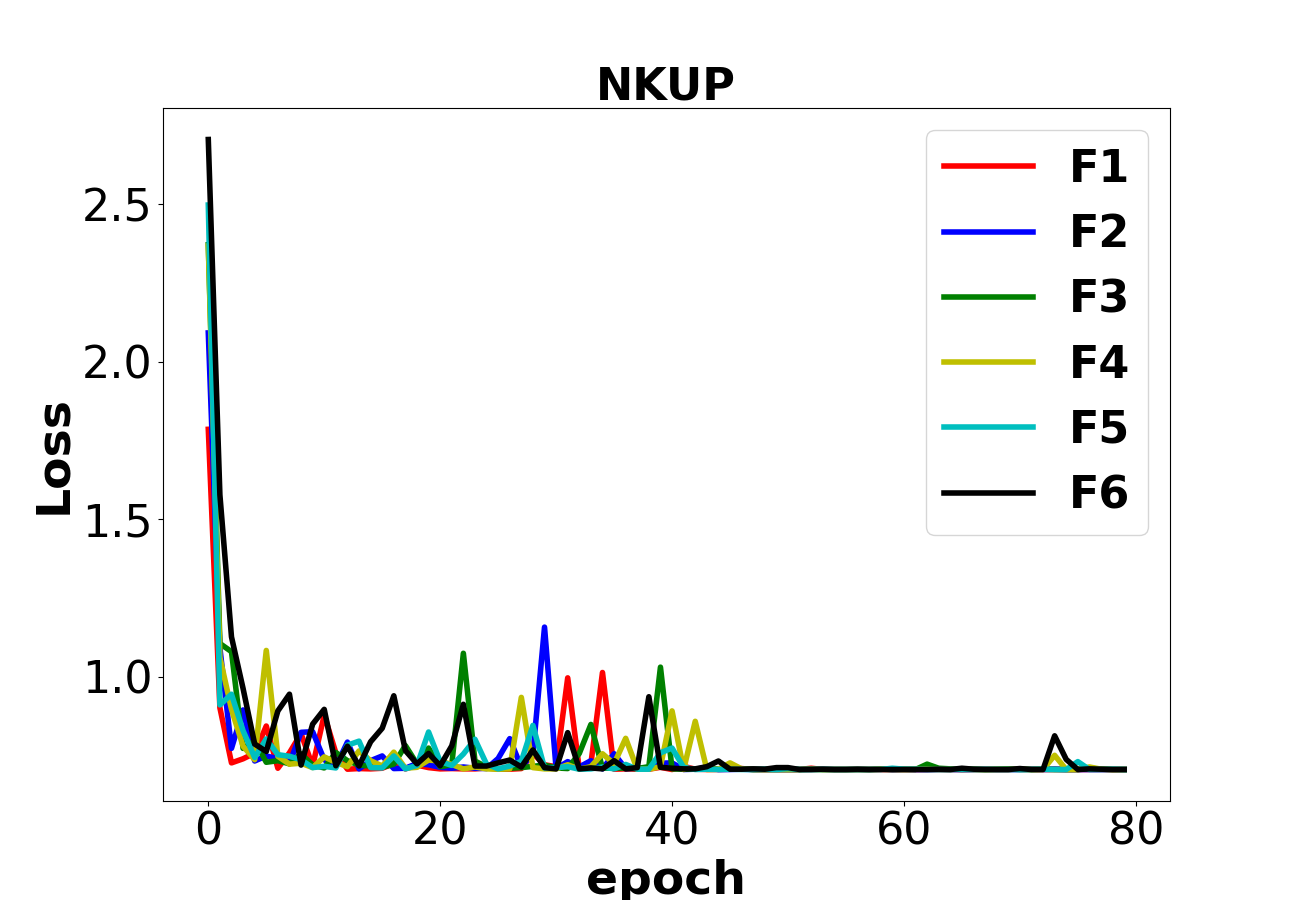}}
\end{minipage}
\caption{Convergence curves of the MVSE method for the LTCC, PRCC, Celeb-reID and NKUP datasets. Note that in the MVSE framework, only a single-granularity feature is partially used. }
\end{figure}

\subsection{Qualitative Visualization}
To further prove the effectiveness and robustness of the MVSE method, in this section, the retrieval results of a) the baseline, b) MGR, c) MGR+CDN, and d) MGR+CDN+PSA (MVSE) on the PRCC and Celeb-reID datasets are given in Fig. 7 and Fig. 8, respectively, where each row is a retrieval example including one query image and the top ten most similar images. Note that the baseline, MGR, MGR+CDN, and MGR+CDN+PSA (or MVSE) have the same meanings as in Table III. The following can be observed from these figures:

1) Due to the settings of the PRCC dataset, only one image in the gallery with the same ID as that of the query image is contained; moreover, the same person in the gallery wears different clothes from those worn in the query image. Therefore, whether the image can be accurately retrieved and whether it is at the forefront of the retrieval results can directly reflect the hit rate of the module. The results are given in Fig. 7. From the figure, we can see that the baseline cannot retrieve the correct results, but when the MGR module is used, the correct results can be returned, thereby proving the effectiveness of the MGR module, but its location is very far back. When the CDN module is further embedded into the baseline, the high-level human attributes can be extracted to effectively describe the person with clothing changes; thus, the correct retrieval results can be returned at the third location. Moreover, when the human visual semantic information is further used to align the human attributes in the PSA module, the performance is further improved, where the location of the correct retrieving results can be returned in position 1. Thus, these experimental results can prove the effectiveness and advantage of the MGR module, CDN module, and PSA module.

2) To better verify the average performances of the modules, including the baseline, MGR, MGR+CDN, and MGR+CDN+PSA (MVSE), we visualize their performances on the Celeb-reID dataset, which contains rich clothing styles and large-scale gallery images with the same ID as that of the query image; the results are shown in Fig. 8. Fig. 8 demonstrates that since the visual appearance of the cloth-changing person drastically changes, the baseline still has difficulty obtaining the correct retrieval results when only the global feature is extracted. When the local features are further combined with the global feature (MGR), the discrimination ability is improved, and one correct result at the ninth location can be returned, but many related person images are ignored. However, when the CDN module is employed, three correct retrieval results at the sixth, eighth and ninth locations can be returned, and then when the PSA module is further utilized, four correct retrieval results can be obtained; moreover, their locations are very close to the front of the images. Thus, these results can further prove that high-level human attributes and visual semantic information help describe cloth-changing people, and the extracted feature is effective and robust.

\begin{figure}[t]
\begin{center}
\includegraphics[width=3.5in,height = 2.0in]{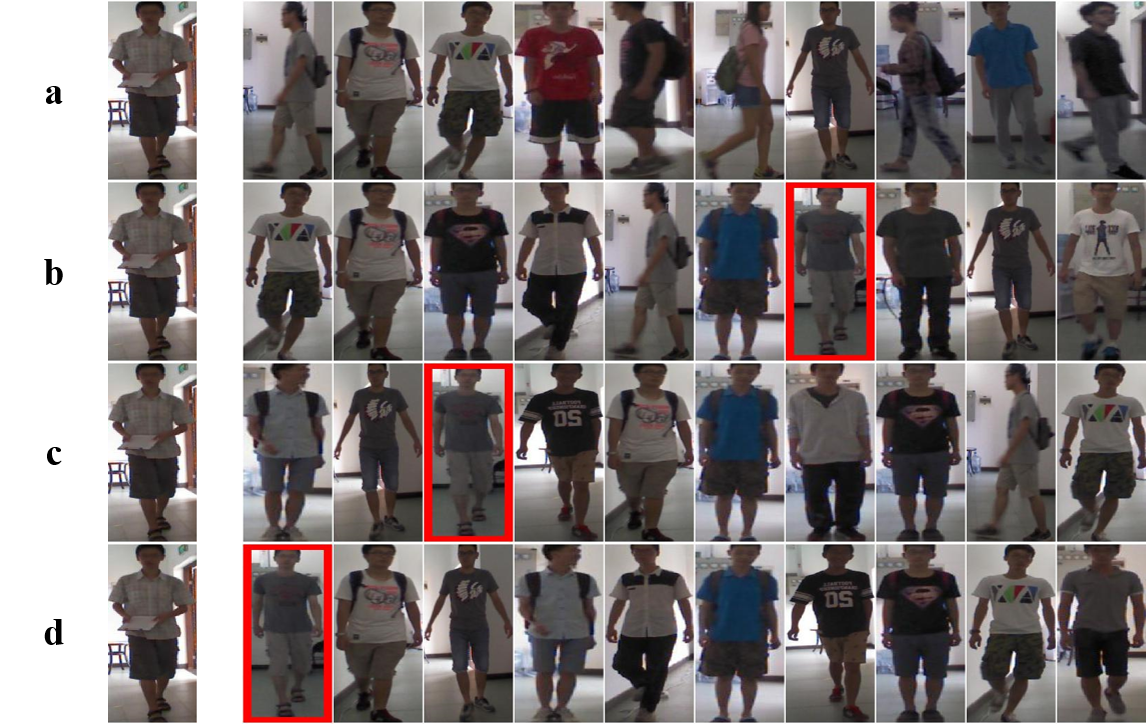}
\caption{Qualitative visualization of a) the baseline, b) MGR, c) MGR+CDN, and d) MGR+CDN+PSA (MVSE) on the PRCC dataset. The top left column is a single query image, and the other columns are the top ten retrieval results. Note that the red box indicates the correct results.}
\end{center}
\end{figure}

\begin{figure}[t]
\begin{center}
\includegraphics[width=3.5in,height = 2.0in]{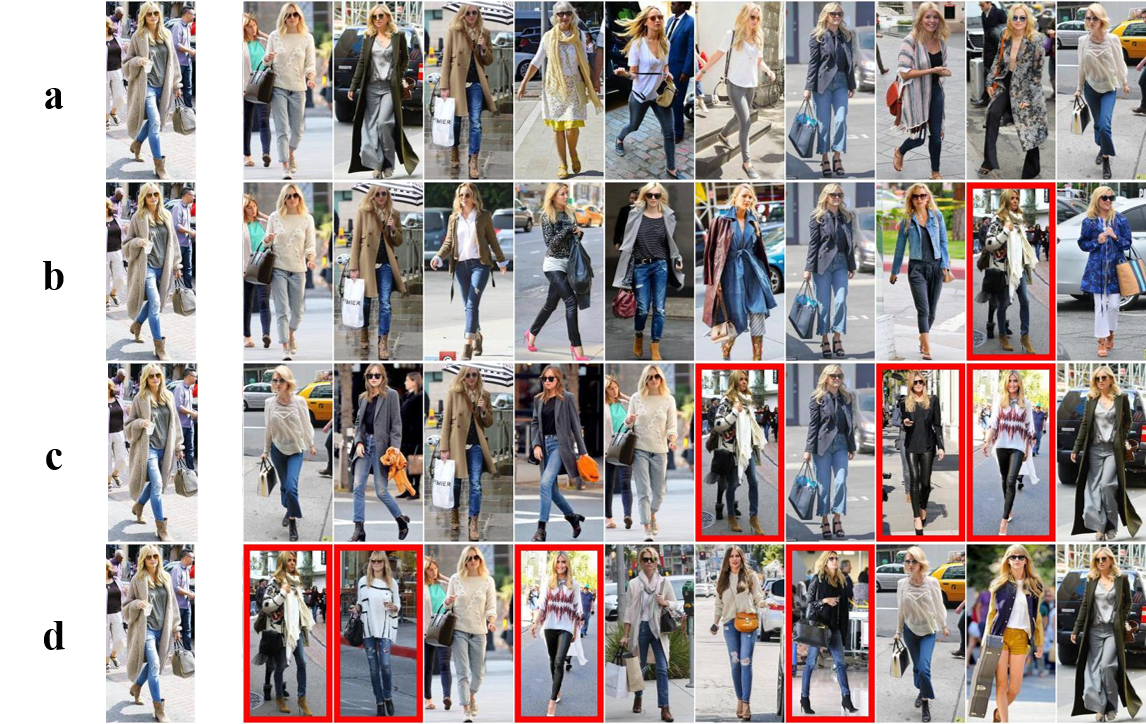}
\caption{Qualitative visualization of a) the baseline, b) MGR, c) MGR+CDN, and d) MGR+CDN+PSA (MVSE) on the Celeb-reID dataset. The top left column is a single query image, and the other columns are the top ten retrieval results. Note that the red box indicates the correct results.}
\end{center}
\end{figure}

\section{Conclusion}

This paper proposes a novel MVSE algorithm for the cloth-changing person ReID task in which feature extraction, the CDN module, and the PSA module are jointly explored in a unified framework. Additionally, we propose an MGR module to focus on the unchanged parts of the human, and then a CDN module is designed to improve the feature robustness of the network for people with different clothes, where different high-level human attributes are fully utilized. Finally, a PSA module is proposed to obtain visual-semantic information, which is used to align the human attributes. The results of extensive experiments conducted on four cloth-changing person ReID datasets demonstrate that MVSE can significantly outperform state-of-the-art cloth-changing person ReID methods in terms of both mAP and rank-1 accuracy. An ablation study also proves that the local features and the global feature help represent cloth-changing persons, and when the MGR module is used, the overall performance can be improved. The CDN module can obtain high-level human attributes that are robust and effective for describing cloth-changing persons. Moreover, visual semantic information can be used to align human attributes, and it can further improve the performance of the proposed method.

In the future, we intend to explore how to shield the effects of clothing changes and design a novel and effective feature fusion module.

\ifCLASSOPTIONcaptionsoff
 \newpage
\fi

 \normalem
 \bibliographystyle{IEEEtran}
 \bibliography{mybib}

\begin{IEEEbiography}[{\includegraphics[width=1in,height=1.25in,clip,keepaspectratio]{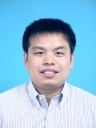}}]{Zan Gao} received his Ph.D degree from Beijing University of Posts and Telecommunications in 2011. He is currently a full Professor with the Shandong Artificial Intelligence Institute, Qilu University of Technology (Shandong Academy of Sciences). From Sep. 2009 to Sep. 2010, he worded in the School of Computer Science, Carnegie Mellon University, USA. From July 2016 to Jan 2017, he worked in the School of Computing of National University of Singapore. His research interests include artificial intelligence, multimedia analysis and retrieval, and machine learning.  He has authored over 80 scientific papers in international conferences and journals including TIP, TNNLS, TMM, TCYBE, TOMM, CVPR, ACM MM, WWW, SIGIR and AAAI, Neural Networks, and Internet of Things.
\end{IEEEbiography}

\begin{IEEEbiography}[{\includegraphics[width=0.8in,height=1in]{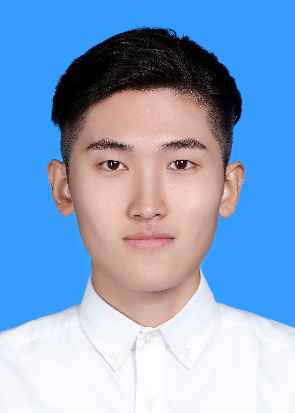}}]{Hongwei Wei} is pursuing his master degree in the Shandong Artificial Intelligence Institute, Qilu University of Technology (Shandong Academy of Sciences). He received his Bachelor degree from Yantao University in 2019. His research interests include artificial intelligence, multimedia analysis and retrieval, computer vision and machine learning
\end{IEEEbiography}

\begin{IEEEbiography}[{\includegraphics[width=0.8in,height=1in,clip,keepaspectratio]{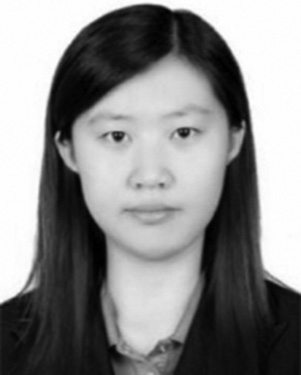}}]{Weili Guan} is now a Ph.D. student with the Faculty of Information Technology, Monash University Clayton Campus, Australia. Her research interests are multimedia computing and information retrieval. She received her bachelor degree from Huaqiao University in 2009. She then obtained her graduate diploma and master degree from National University of Singapore in 2011 and 2014 respectively. After that, she joined Hewlett Packard enterprise Singapore as software engineer and worked there for around five years. 
\end{IEEEbiography}

\begin{IEEEbiography}[{\includegraphics[width=0.8in,height=1in,clip,keepaspectratio]{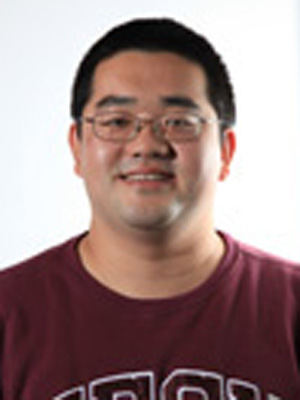}}]{An-An Liu} is a full professor in the school of electronic engineering, Tianjin University, China. He received his Ph.D degree from Tianjin University in 2010. His research interests include computer vision and machine learning.
\end{IEEEbiography}

\begin{IEEEbiography}[{\includegraphics[width=0.8in,clip,keepaspectratio]{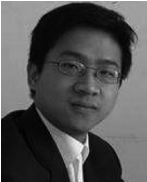}}]{Meng Wang} received the BE and PhD degrees in the Special Class for the Gifted Young and the Department of Electronic Engineering and Information Science from the University of Science and Technology of China (USTC), Hefei, China, respectively. He is a professor at the Hefei University of Technology, China. His current research interests include multimedia content analysis, search, mining, recommendation, and large-scale computing. He received the best paper awards successively from the 17th and 18th ACM International Conference on Multimedia, the best paper award from the 16th International Multimedia Modeling Conference, the best paper award from the 4th International Conference on Internet Multimedia Computing and Service, and the best demo award from the 20th ACM International Conference on Multimedia.
\end{IEEEbiography} 




\end{document}